%% file: acl_latex.tex
\pdfoutput=1
\documentclass[11pt]{article}

\usepackage[final]{acl}

\usepackage{times}
\usepackage{latexsym}

\usepackage[T1]{fontenc}
\usepackage[utf8]{inputenc}

\usepackage{microtype}

\usepackage{inconsolata}

\usepackage{graphicx}

\usepackage{amsmath}
\usepackage{wrapfig}
\usepackage{graphicx}
\usepackage{subcaption}
\usepackage{xcolor}
\usepackage{adjustbox}
\usepackage{diagbox}
\usepackage{booktabs}
\usepackage{multirow}
\usepackage[most]{tcolorbox}
\tcbset{
  colback=gray!5,
  colframe=gray!80,
  boxrule=0.4pt,
  arc=4pt,
  outer arc=4pt,
  boxsep=5pt,
  left=6pt,
  right=6pt,
  top=6pt,
  bottom=6pt,
  enhanced
}

\title{Out of Style: RAG’s Fragility to Linguistic Variation}

\author{Tianyu Cao$^1$\thanks{Equal contribution.} \quad Neel Bhandari$^{1*}$ \quad Akhila Yerukola$^1$ \quad Akari Asai$^{12}$ \quad Maarten Sap$^1$ \\
$^1$ Language Technologies Institute, Carnegie Mellon University \\
$^2$ Allen Institute for AI \\
\texttt{\{tianyuca, neelbhan, ayerukol, aasai, msap2\}@cs.cmu.edu}
}

\begin{document}
\maketitle

\input{sections/00_Abstract}

\input{sections/01_Introduction}
\input{sections/02_Related_Work}
\input{sections/03_Robustness_Evaluation_Approach}
\input{sections/04_Experimental_Setups}
\input{sections/05_Robustness_Experimental_Results}

\input{sections/06_Conclusion}

% \clearpage
\section*{Limitations}

% \maarten{Some ideas (but please also brainstorm):
% - Only 4 styles (5 rewrites) -- while it's representative of a broad range of variation (stylistic, pragmatic, structural), there are other variations that could be explored in future work. Give some examples.
% - Mitigation: Trivial solution is to train a model with automatically rewritten queries, which future work should explore. 
% - Only two rewriting models; robustness of RAG could potentially be different if rewrites were done with another model, though we expect that the same patterns will appear given our results.
% }
While our choice of linguistic dimensions cover a broad spectrum of stylistic, pragmatic, and structural variations, other relevant factors such as dialect, idiomatic expressions, or domain-specific terminology could be explored in future work. 
Our grammatical variations (typos and round-trip translation) represent a subset of potential errors; other categories such as verb form or agreement errors~\cite{dahlmeier2013building} may present greater challenges and warrant future investigation.
We conducted query rewriting using two LLMs (GPT-4o-mini and Llama-3.1-70B-Instruct) and observed similar vulnerabilities; future studies may verify the generalizability of these findings using a broader range of rewriting methods.
Additionally, while we explored widely used methods such as query expansion and reranking to test for mitigation strategies, more comprehensive approaches, including training models explicitly on diverse, linguistically varied data, remain important avenues for future research.

\section*{Acknowledgments}
This research was supported in part by the National Science Foundation under grant 2230466 and in part by DSO National Laboratories.

% \nocite{*}
\clearpage
\bibliography{custom}

\clearpage
\appendix
\input{sections/07_Appendix}

\end{document}

%% file: sections/00_Abstract.tex
\begin{abstract} 

Despite the impressive performance of Retrieval-augmented Generation (RAG) systems across various NLP benchmarks, their robustness in handling real-world user-LLM interaction queries remains largely underexplored.
This presents a critical gap for practical deployment, where user queries exhibit greater linguistic variations and can trigger cascading errors across interdependent RAG components.
In this work, we systematically analyze how varying four linguistic dimensions 
({\it formality, readability, politeness, and grammatical correctness}) 
impact RAG performance.
We evaluate two retrieval models and nine LLMs, ranging from 3 to 72 billion parameters, across four information-seeking Question Answering (QA) datasets.
Our results reveal that linguistic reformulations significantly impact both retrieval and generation stages, leading to a relative performance drop of up to 40.41\% in Recall@5 scores for less formal queries and 38.86\% in answer match scores for queries containing grammatical errors. 
Notably, RAG systems exhibit greater sensitivity to such variations compared to LLM-only generations, highlighting their vulnerability to error propagation due to linguistic shifts. 
These findings highlight the need for improved robustness techniques to enhance reliability in diverse user interactions.
\footnote{Code is available at 
\url{https://github.com/Springcty/RAG-fragility-to-linguistic-variation}.}
\looseness=-1
\end{abstract}

%% file: sections/01_Introduction.tex
\section{Introduction}

\begin{figure*}
    \centering
    \includegraphics[width=0.9\linewidth]{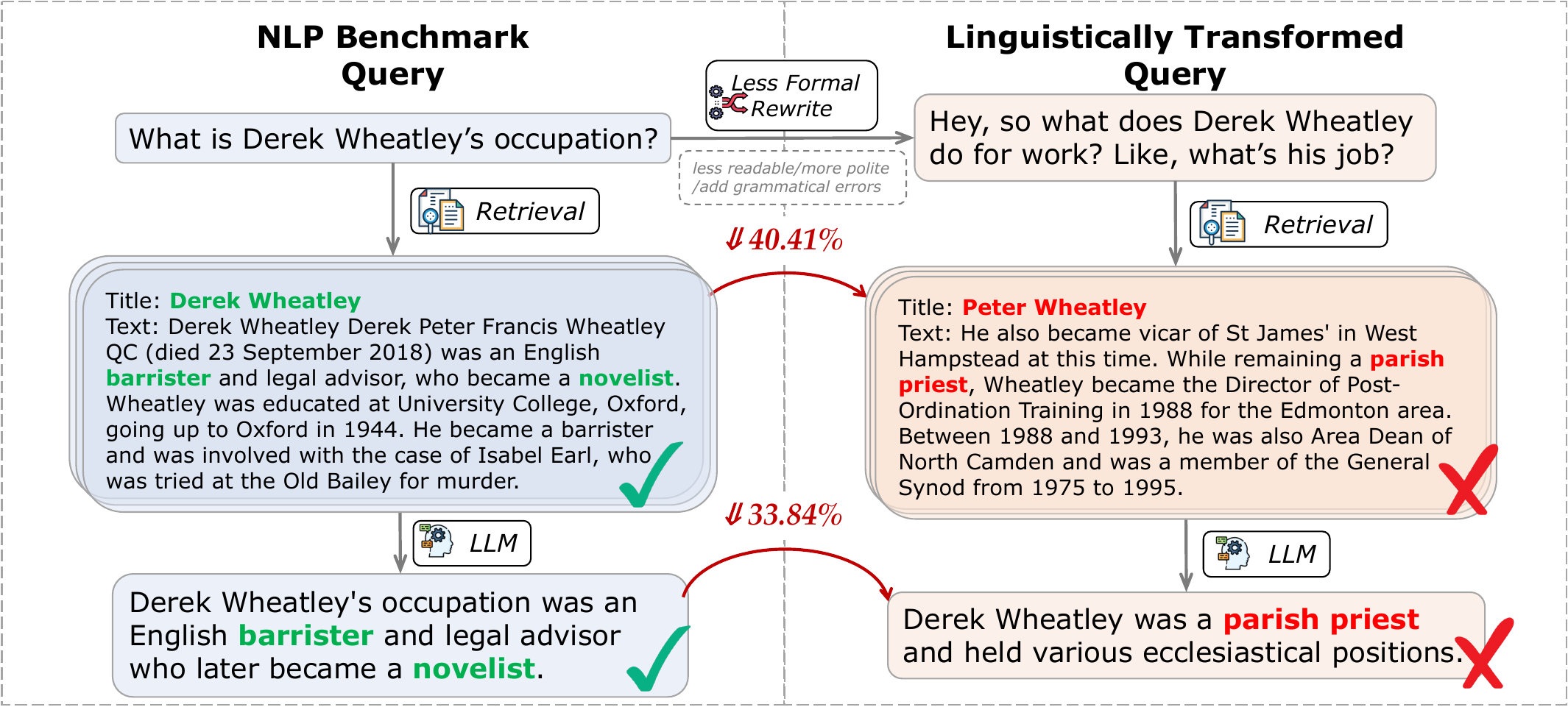}
    \caption{
    \textbf{RAG systems demonstrate overall performance degradation when queries are rewritten to be less formal, more polite, less readable, and with grammatical errors.}
    For the traditional NLP query (left), the RAG systems successfully retrieve related information and generate the correct answer, while the less formal queries (right) retrieve incorrect information.
    The linguistic variation on formality causes significant performance drops: 40.41\% decrease in Recall@5 and 33.38\% decrease in answer match (AM) score on the MS MARCO dataset.
    }
    \label{fig:teaser}
\end{figure*}

Retrieval-augmented Generation (RAG) systems enhance Large Language Models (LLMs) by integrating external knowledge retrieval, grounding their output in factual context to improve accuracy and reliability~\citep{lewis2021retrievalaugmentedgenerationknowledgeintensivenlp, gao2024retrievalaugmentedgenerationlargelanguage}. 
However, their widespread integration into real-world applications~\citep{K2viewGenAI2024} introduces potential challenges regarding robustness to linguistic variations.
Users bring varied backgrounds, domains, and cultural contexts that naturally produce linguistic differences in their queries~\citep{park2024valuescopeunveilingimplicitnorms, li2020studying, lorenzo2013cross}. As Figure \ref{fig:teaser} illustrates, different from the carefully curated queries from traditional NLP benchmarks, real-world user-LLM queries tend to be less formal and frequently contain grammatical inconsistencies~\citep{ouyang2023shiftedoverlookedtaskorientedinvestigation}.

Failing to account for these linguistic variations risks excluding a broad segment of users from effective interaction, especially for users whose linguistic expressions fall outside the narrow patterns these systems are tuned on \citep{liang2023gptdetectorsbiasednonnative}. 
Moreover, unlike standalone LLMs, RAG systems incorporate multiple interdependent components, making them susceptible to cascading errors arising at both the retrieval and generation stages~\citep{asai2023selfraglearningretrievegenerate,yoran2024making,kim2025mitigating}.
{\it A truly robust RAG system should maintain consistent retrieval effectiveness and generation quality across the full spectrum of user linguistic variations.}
\looseness=-1

We present a large-scale systematic investigation of how variations in linguistic characteristics affect the robustness of RAG systems.
% add explanation on linguistic choices
We target diverse and prevalent variations commonly found in real-world user inputs that meaningfully challenge RAG systems~\citep{park2024valuescopeunveilingimplicitnorms, ouyang2023shiftedoverlookedtaskorientedinvestigation}, namely, 
\textbf{formality}, \textbf{readability}, \textbf{politeness}, and \textbf{grammatical correctness}.
These choices ensure our analysis covers stylistic, pragmatic, and structural aspects aligned with practical usage.
By automatically rewriting queries across these dimensions, we analyze how linguistic variations impact each RAG system component, as well as potential cascading errors throughout the pipeline.
Our evaluation encompasses two retrieval models, namely Contriever~\citep{izacard2021contriever} and ModernBERT~\citep{nussbaum2024nomic}, and nine LLMs from three families (Llama 3.1, Qwen 2.5, Gemma 2) of varying scales across four open-domain Question Answering (QA) datasets: PopQA~\citep{mallen2023trustlanguagemodelsinvestigating}, EntityQuestions~\citep{sciavolino2022simpleentitycentricquestionschallenge}, MS MARCO~\citep{bajaj2018msmarcohumangenerated}, and Natural Questions~\citep{kwiatkowski-etal-2019-natural}.
\looseness=-1

Our experiments reveal significant vulnerabilities in RAG systems to linguistic variations.
Retrieval analysis shows an average 15\% relative performance degradation across all datasets and linguistic dimensions, with grammatical modifications most severely impacting recall while politeness variations show minimal effect.
Generation analysis demonstrates average decreases of 16.52\% (AM), 41.15\% (EM), and 19.60\% (F1) across all experimental settings.
Notably, increasing the LLM scales doesn't always help mitigate these performance gaps.

Furthermore, RAG systems exhibit greater vulnerability to linguistic variations than LLM-only generations, suggesting cascading errors across components. 
The PopQA dataset shows a 22.53\% average performance drop in RAG systems versus only 10.78\% in LLM-only generations. 
These findings highlight the urgent need to improve the robustness of retrieval components when handling linguistically varied queries.
We also find that while advanced RAG techniques, e.g., query expansion with HyDE~\citep{gao2022precisezeroshotdenseretrieval} and documents reranking, tend to improve overall performance, they remain similarly vulnerable to linguistic variations.
\looseness=-1

In summary, we present the first systematic analysis of the robustness of RAG systems to linguistic query reformulation.
Our results demonstrate that despite strong performance on standard benchmarks, they remain fragile to inevitable real-world linguistic variations. 
These findings highlight the need for enhanced robustness techniques to improve reliability across diverse user interactions and inform design principles for next-generation RAG systems.

%% file: sections/02_Related_Work.tex
\section{Related Work} 

\paragraph{Robustness of retrieval systems to linguistic variations.}
Prior research investigated retrieval system robustness to noisy queries \citep{campos2023noiserobustdenseretrievalcontrastive, chen2022towards, chen2023dealing} and specific linguistic variations including word substitutions \citep{Wu_2022}, aspect changes and paraphrasing \citep{penha2022evaluatingrobustnessretrievalpipelines}, typographical errors \citep{zhuang2021dealingtyposbertbasedpassage}, and grammatical variations \citep{long2024whispersgrammarsinjectingcovert}. 
Our work provides the first holistic evaluation of RAG systems’ robustness to diverse linguistic variations and uncovers cascading failures across the entire RAG pipeline. 

\paragraph{Robustness of language models to linguistic variations.}
Prior research has examined impacts of syntactic perturbations \cite{moradi-samwald-2021-evaluating, singh2024robustnessllmsperturbationstext}, round-trip translation \cite{Bhandari_2023}, politeness variations \cite{yin2024respectllmscrosslingualstudy}, equivalent queries \cite{cao2024worstpromptperformancelarge} and scale of model size on robustness to grammatical errors \citep{yin-etal-2020-robustness, hagen-etal-2024-revisiting}. \citet{rawte2023exploringrelationshipllmhallucinations} examined how formality and readability affect LLM performance in isolation. In contrast, our research investigates a broader spectrum of linguistic variations, generates data for each variation using LLM-based rewrites, and studies their compounding effects throughout the end-to-end RAG pipeline comprehensively.

\paragraph{Robustness of RAG systems.}
While RAG systems demonstrate impressive performance \citep{lewis2021retrievalaugmentedgenerationknowledgeintensivenlp, gao2024retrievalaugmentedgenerationlargelanguage} and reduced hallucinations \citep{mallen2023trustlanguagemodelsinvestigating}, vulnerabilities exist with increasing retrieval context noise \citep{chen2023benchmarkinglargelanguagemodels}, irrelevant contexts \citep{yoran2024makingretrievalaugmentedlanguagemodels}, and noise impacts \citep{fang2024enhancingnoiserobustnessretrievalaugmented}. \citet{yang2025quantifyingrobustnessretrievalaugmentedlanguage} analyzed how spurious features affect RAG perform. 
\citet{cho2024typosbrokeragsback} introduced document-level perturbations and evaluated RAG's vulnerability to noisy documents.
RbFT~\citep{10.1145/3726302.3730078} introduce specialized training objectives to improve consistency under corrupted or adversarial retrieval contexts.
These studies focus primarily on retrieved content noise rather than initial \textit{query}; our work uniquely demonstrates how diverse linguistic variations in user queries compound throughout the RAG pipeline, exposing critical vulnerabilities in systems serving diverse users.

%% file: sections/03_Robustness_Evaluation_Approach.tex
\section{Robustness Evaluation Approach}

In our work, we explore the impact of the following linguistic aspects: {\bf Formality}, {\bf Readability}, {\bf Politeness} and {\bf Grammatical Correctness - Round-Trip Translation and Typos}. We explore these linguistic queries as they are essential dimensions of language variation that are prevalent and significant in real-world RAG interactions.
We extend \citet{rawte2023exploringrelationshipllmhallucinations}'s findings on linguistic variations and LLM hallucinations by synthetically generating queries across four linguistic dimensions and four datasets to analyze each RAG pipeline component comprehensively. We first formulate our task, followed by defining each of our linguistic characteristics (Section~\ref{sec:linguistic_variations}), and elaborating on our query rewriting design~(Section~\ref{subsec:query-rewrite}).

\paragraph{Task formulation.}
% \akari{you might make it to a paragraph, not subsection.}
Given a seed dataset $\mathcal{D} = \{(x_1, y_1), (x_2, y_2), \ldots\}$, where $x_i$ and $y_i$ indicate $i$-th input and output, respectively, we reformulate each query $x_i \rightarrow x_i'$ based on four linguistic aspects. A robust RAG system, composed of a retriever $\mathcal{R}$ and a generator $\mathcal{G}$ operating on corpus $\mathcal{C}$, should maintain performance when processing linguistically varied inputs that are semantically equivalent, i.e., where the gold output remains unchanged semantically. For the retrieval component, we expect retrieved documents $\mathbf{D}_i = \mathcal{R}(x_i, \mathcal{C})$ and $\mathbf{D}_i' = \mathcal{R}(x_i', \mathcal{C})$ to both contain the answer $y_i$. For generation, given retrieved documents $\mathbf{D}_i, \mathbf{D}_i'$, a robust system should produce: $\mathcal{G}(x_i, \mathbf{D}_i) \approx \mathcal{G}(x_i', \mathbf{D}_i') \approx y_i$.

\subsection{Linguistic Variations}
\label{sec:linguistic_variations}

\paragraph{Formality.}
Formality in language lacks universal definition \citep{pavlick-tetreault-2016-empirical,Mosquera2011TheUO,fang-cao-2009-adjective}, but encompasses situational factors \citep{hovy1987generating,Lahiri2011InformalityJA}, grammar quality \citep{peterson-etal-2011-email}, and specific linguistic elements like contractions \citep{heylighen1999formality}. We quantify formality using the RoBERTa-based formality classifier from \citep{formality_model}. 

\paragraph{Readability.}
Readability quantifies text comprehensibility through linguistic complexity. We employ the widely-used Flesch Reading Ease Score (FRES; \citealt{Flesch1948ANR}) to assess readability \citep{rawte2023exploringrelationshipllmhallucinations, han2024usereadabilitymetricslegal}, defined in \ref{sec:app_ling}.

\paragraph{Politeness.} 
Politeness is a sociocultural phenomenon defined as showing consideration of others~\citep{wang2014model}.
We calculate politeness scores using Polite Guard~\citep{politeguard}, an open-source NLP model from Intel that's fine-tuned from BERT to classify text into four levels: \textit{polite}, \textit{somewhat polite}, \textit{neutral}, and \textit{impolite}.
% For our query rewriting purposes, we consider both the \textit{polite} and \textit{somewhat polite} classifications as successful outcomes when transforming original queries to more polite versions.

\paragraph{Grammatical correctness.}
\label{para:grammatical}
In our work, we define grammatical correctness as the preservation of both grammaticality \citep{chomsky2002syntactic} and semantic fidelity.
We alter the grammatical correctness through two approaches, inspired by \citet{yin-etal-2020-robustness, zhuang2021dealingtyposbertbasedpassage, lichtarge-etal-2019-corpora}: (1) \textbf{Typos}, where random addition, deletion, or substitution operations are applied at a 20\% probability per word, requiring an edit distance of at least 1; and (2) \textbf{Round-trip translation (RTT)} via English-Afrikaans-English using EasyNMT with opus-mt model, requiring the output to not be the exact same as the original query.

\subsection{Query Rewriting}
\label{subsec:query-rewrite}

\begin{table*}[t]
\centering
% \small
\resizebox{\textwidth}{!}{%
\begin{tabular}{@{}l p{0.9\textwidth}@{}}
\toprule
\textbf{Linguistic Dimension}  & \textbf{Example Rewrites} \\
\textbf{(Dataset)} & \\
\midrule
\textbf{Politeness $\uparrow$} 
  & \textbf{Original:} complex carbohydrates are stored in animals in the form of\\
  (MS MARCO)& \textbf{Rewritten:} Would you be so kind as to share how complex carbohydrates are stored in animals?\\
\midrule
\textbf{Readability $\downarrow$}
  & \textbf{Original:} who stars in the new movie the post\\
  (Natural Questions) & \textbf{Rewritten:} In the upcoming cinematic production titled ``The Post,'' which individuals have been cast in leading roles?\\
\midrule
\textbf{Formality $\downarrow$}
  & \textbf{Original:} Who is the author of Dolores Claiborne?\\
  (PopQA) & \textbf{Rewritten:} Hey, do you happen to know who wrote Dolores Claiborne? I'm kinda curious!\\
\midrule
\textbf{Grammar: RTT $\downarrow$}
  & \textbf{Original:} Which company is HMS Blankney produced by?\\ (EntityQuestions)
  & \textbf{Rewritten:} What company is producing HMS Blankey?\\
\midrule

\textbf{Grammar: Typos $\downarrow$} 
  & \textbf{Original:} when did the japanese river otter \textbf{become} extinct \\ 
  (Natural Questions)
  & \textbf{Rewritten:} when did the japanese river otter \textbf{ecome} extinct\\
\bottomrule
\end{tabular}%
}
\caption{Examples of query rewrites across different linguistic dimensions and datasets. Each pair shows the original query and its rewritten form.}
\label{tab:rewrite_examples}
\end{table*}

We systematically reformulate queries $x_i \rightarrow x_i'$ across targeted linguistic dimensions while ensuring all rewrites satisfy dimension-specific thresholds and preserve semantic consistency (Appendix \ref{sec:app_ling}). 

\paragraph{Linguistic reformulation.}
For each distinct dimension (formality, readability, and politeness), we sample $5,000$ original queries and use GPT-4o-mini \citep{openai2024gpt4technicalreport} to generate rewrites using three different prompts (detailed in Appendix \ref{app: rewriting_prompts}).\footnote{To verify that our experimental results are robust to rewriting models, we generate $500$ rewritten queries from the PopQA dataset using Llama-3.1-70B-Instruct. Shown in Appendix~\ref{app:llama_rewriting}, results confirm our findings with GPT-4o-mini.}
We design distinct query sets across all datasets to be less formal, less readable, or more polite—directions chosen because the well-curated original datasets contained predominantly formal, readable, and neutral language, allowing our modifications to explore more realistic linguistic variations. 
This creates $15,000$ rewritten samples per dimension for each dataset,
over which we average results to account for prompt stochasticity. 
For grammatical correctness, we similarly use $5,000$ original queries but apply deterministic transformations through typo introduction and round-trip translation as described in Section~\ref{para:grammatical}, creating $5,000$ modified samples for each approach.
Example rewrites are shown in Table~\ref{tab:rewrite_examples}.

\paragraph{Semantic consistency.}
We define a rewritten query $x_i'$ as semantically consistent with the original query $x_i$ if it preserves the original pragmatic intent, i.e., a human annotator would assign the same gold answer $y_i$ to both queries.
This operational definition reflects the requirements that linguistic variation should not alter the underlying information need.
To strictly enforce this, we apply a two-stage verification procedure.
First, we automatically filter rewrites using a sentence-level semantic similarity threshold ($>0.7$) computed with MPNet-v2~\citep{song2020mpnetmaskedpermutedpretraining}.
Second, we conduct qualitative evaluation through manual annotation on $250$ queries across all linguistic variations, finding that 94.67\% of the rewritten queries preserve the original answer. Further details of the annotation are provided in Appendix~\ref{sec:semantic_preservation}.

For each linguistic dimension and dataset, we construct controlled comparative datasets $\mathcal{D}' = \{(x_1', y_1), (x_2', y_2), \ldots\}$, enabling direct measurement of performance changes attributable solely to linguistic variation. 

% To validate our approach preserves semantic consistency, we annotate $250$ queries across all linguistic variations, finding that 94.67\% of the rewritten queries preserve the semantic meaning of the original query\CR{, i.e., the original answer is still the right answer}. Details are available in Appendix~\ref{sec:semantic_preservation}.
% \looseness=-1

%% file: sections/04_Experimental_Setups.tex
\section{Experimental Setups}

\subsection{Benchmarks}
In this work, we use four open-domain QA datasets as seed datasets $\mathcal{D}$ and evaluate the effects how linguistic variations of the original queries affect RAG systems. 
PopQA~\citep{mallen2023trustlanguagemodelsinvestigating} is a large-scale entity-centric open-domain QA dataset about entities with a wide variety of popularity. 
EntityQuestions~\citep{sciavolino2022simpleentitycentricquestionschallenge} is a set of simple, entity-rich questions based on facts of Wikidata.
MS MARCO~\citep{bajaj2018msmarcohumangenerated} contains questions derived from real user search queries from Bing's search logs.
Natural Questions~\citep{kwiatkowski-etal-2019-natural} contains questions consisting of real, anonymized, aggregated queries to the Google search engine.
Both PopQA and EntityQuestions consist of well-structured, standardized, and simple queries, while queries from MS MARCO and Natural Questions exhibit free-form and arbitrary.
We evaluate using PopQA's test split, EntityQuestions' dev split, and both dev and test splits from MS MARCO and Natural Questions.
The retrieval is performed on the Wikipedia passage set used in DPR\footnote{\url{https://dl.fbaipublicfiles.com/dpr/wikipedia_split/psgs_w100.tsv.gz}} for the PopQA, EntityQuestions, and Natural Questions datasets, while the MS MARCO dataset uses its corresponding passage dataset~\citep{bajaj2018msmarcohumangenerated}.

\subsection{Models}
\paragraph{Retrieval.}
We use two neural retrieval systems, namely \textbf{Contriever}~(\texttt{facebook/contriever}; ~\citealt{izacard2022unsuperviseddenseinformationretrieval}) and \textbf{ModernBERT Embed}~(\texttt{nomic-ai/modernbert-embed-base};~\citealt{modernbert, nussbaum2024nomic}).
Contriever is an unsupervised dense retriever built from BERT base architecture and is pre-trained using a contrastive learning framework.
ModernBERT Embed is an embedding model trained from ModernBERT-base~\citep{modernbert}, bringing the new advances of ModernBERT to embeddings.
We use the retrieval system implementation by ~\citet{izacard2022unsuperviseddenseinformationretrieval} for both retrieval models.\footnote{\url{https://github.com/facebookresearch/contriever}} 

\paragraph{Generation.}
We evaluate the nine most advanced open-source instruction-tuned LLMs from three model families with various scales:
\textbf{Llama 3.1} (\citealt{grattafiori2024llama3herdmodels}; 8 and 70 billion),
\textbf{Qwen 2.5} (~\citealt{qwen2, qwen2.5}; 3, 7, 32, and 72 billion),
and \textbf{Gemma 2} (~\citealt{gemmateam2024gemma2improvingopen}; 2, 9, and 27 billion).
Detailed hyperparameter settings are included in the Appendix~\ref{app:gen_setting}.

We use few-shot prompting to ensure that the model outputs are in the correct format.
For each dataset and linguistic characteristic, we include two question-answer pairs in the context: one random original query with its answer, and its corresponding linguistically rewritten version with the same answer.
This balanced approach exposes the model to both original and rewritten query formats to ensure fairness.
We also include the top five retrieved passages in the context.
Detailed prompts are provided in Appendix~\ref{app:llm_prompts}.

\subsection{Metrics}

\paragraph{Retrieval.} We employ \textbf{Recall@k (R@k)}~\citep{karpukhin2020densepassageretrievalopendomain}, which calculates the fraction of the retrieved documents containing gold answers. We use $k=5$ as our primary setup, and evaluate the effect of varying $k$ in our analysis.

\paragraph{Generation.} The generation stage is assessed using a comprehensive set of metrics: \textbf{Answer Match (AM)} measures the percentage of the predictions that any substring of the prediction is an exact match of any of the ground truth answers, \textbf{Exact Match (EM)} measures the percentage of predictions that exactly match any of the ground truth answers, and \textbf{F1 Score (F1)} captures the harmonic mean of precision and recall in generated responses. 
We mainly report AM scores in the main paper because it better reflects answer correctness under linguistic and stylistic variaion. The full results could be found in Appendix~\ref{app:full_rag_results}.
As a complement to standard metrics, we also use \textbf{LLM-as-a-Judge} with GPT-5-mini~\citep{openai_gpt5_system_card}.

%% file: sections/05_Robustness_Experimental_Results.tex
\section{RAG Robustness Experimental Results}

In this section, we progress from component-level to end-to-end RAG system analysis, followed by an assessment of advanced techniques (query expansion and re-ranking) and their ability to mitigate performance drops when faced with linguistic variations.

\subsection{Retrieval Analysis}
\label{sec:retrieval}
We conduct a comprehensive analysis of retrieval systems performance across two candidate retrievers: Contriever and ModernBERT Embed. The results are shown in Table~\ref{tab:retriever_performance}.

\begin{table}[th]
\centering
\setlength{\tabcolsep}{2pt}
\small
\resizebox{\columnwidth}{!}{%
    \begin{tabular}{llcccc c}
\toprule
\textbf{Linguistics} & \textbf{Retriever} & \textbf{PopQA} & \textbf{Entity} & \textbf{MARCO} & \textbf{NQ} & \textbf{$\Delta$ Q-len}\\
\midrule

\multirow{2}{*}{Readability} 
    & Contriever & 18.45 \tiny{(\textcolor{gray}{0.61})} & 8.61 \tiny{(\textcolor{gray}{0.64})} & \textbf{21.10} \tiny{(\textcolor{gray}{0.25})} & 7.69 \tiny{(\textcolor{gray}{0.60})} & \multirow{2}{*}{5.23} \\
    & ModernBERT & \textbf{17.73} \tiny{(\textcolor{gray}{0.65})} & 13.17 \tiny{(\textcolor{gray}{0.61})} & 14.58 \tiny{(\textcolor{gray}{0.40})} & 10.08 \tiny{(\textcolor{gray}{0.65})} & \\
\midrule

\multirow{2}{*}{Gram. (RTT)} 
    & Contriever & \textbf{29.00} \tiny{(\textcolor{gray}{0.59})} & 14.57 \tiny{(\textcolor{gray}{0.68})} & 9.14 \tiny{(\textcolor{gray}{0.34})} & 23.50 \tiny{(\textcolor{gray}{0.62})} & \multirow{2}{*}{-0.26} \\
    & ModernBERT & \textbf{29.68} \tiny{(\textcolor{gray}{0.62})} & 14.85 \tiny{(\textcolor{gray}{0.66})} & 17.54 \tiny{(\textcolor{gray}{0.32})} & 18.24 \tiny{(\textcolor{gray}{0.67})} & \\
\midrule

\multirow{2}{*}{Gram. (Typos)} 
    & Contriever & 27.79 \tiny{(\textcolor{gray}{0.59})} & 14.80 \tiny{(\textcolor{gray}{0.68})} & 15.53 \tiny{(\textcolor{gray}{0.34})} & \textbf{30.83} \tiny{(\textcolor{gray}{0.62})} & \multirow{2}{*}{0.02}\\
    & ModernBERT & \textbf{22.48} \tiny{(\textcolor{gray}{0.62})} & 11.01 \tiny{(\textcolor{gray}{0.66})} & 12.30 \tiny{(\textcolor{gray}{0.32})} & 13.45 \tiny{(\textcolor{gray}{0.67})} & \\
\midrule

\multirow{2}{*}{Formality} 
    & Contriever & 19.96 \tiny{(\textcolor{gray}{0.70})} & 10.71 \tiny{(\textcolor{gray}{0.68})} & \textbf{40.41} \tiny{(\textcolor{gray}{0.25})} & 15.35 \tiny{(\textcolor{gray}{0.65})} & \multirow{2}{*}{13.65}\\
    & ModernBERT & 13.67 \tiny{(\textcolor{gray}{0.74})} & 8.05 \tiny{(\textcolor{gray}{0.68})} & \textbf{15.55} \tiny{(\textcolor{gray}{0.40})} & 9.51 \tiny{(\textcolor{gray}{0.69})} & \\
\midrule

\multirow{2}{*}{Politeness} 
    & Contriever & 8.30 \tiny{(\textcolor{gray}{0.62})} & 1.70 \tiny{(\textcolor{gray}{0.67})} & \textbf{16.44} \tiny{(\textcolor{gray}{0.26})} & 1.16 \tiny{(\textcolor{gray}{0.60})} & \multirow{2}{*}{7.29}\\
    & ModernBERT & \textbf{10.70} \tiny{(\textcolor{gray}{0.67})} & 3.39 \tiny{(\textcolor{gray}{0.67})} & 5.18 \tiny{(\textcolor{gray}{0.40})} & 4.97 \tiny{(\textcolor{gray}{0.65})} &\\
\bottomrule
\end{tabular}
}
\caption{
Relative retrieval performance drop (\%) in R@5 on rewritten queries across datasets.
(\textcolor{gray}{Original scores}) shown in gray parentheses. Bold indicates \textbf{the largest degradation value} per retriever. 
\textbf{$\Delta$ Q-len} represents average query token-length change.
\textbf{Query linguistic variations degrades retrieval performance consistently across all linguistic characteristics.}
}
\label{tab:retriever_performance}
\end{table}

\paragraph{Query variations based on linguistic dimensions degrade retrieval performance.}
Our analysis (Table \ref{tab:retriever_performance}) reveals significant linguistic fragility in retrieval systems, with performance degradation averaging 16.7\% for Contriever and 13.3\% for ModernBERT across all modifications. 
We hypothesize that ModernBERT’s increased robustness could largely be attributed to its diverse pre-training data, which forces the model to learn to process non-standard sequences and enhances its resilience to linguistic noise. Additionally, the masked language modeling objective utilized by ModernBERT could effectively train the model to denoise and reconstruct context from corrupted inputs. 
Results show highest sensitivity on PopQA (19.78\% average impact), particularly to grammatical transformations (29.34\% from RTT). MS MARCO exhibits the second-highest impact (16.78\%), with striking sensitivity to formality changes (40.41\% with Contriever), suggesting retrieval systems may be implicitly optimized for specific linguistic patterns, limiting effectiveness when handling diverse query variations.

\begin{table*}[!ht]
\centering

\begin{subtable}{\textwidth}
\setlength{\tabcolsep}{4pt}
\resizebox{\textwidth}{!}{
\begin{tabular}{l cccc cccc cccc}\toprule
 \multirow{3}{*}{%
        \diagbox[width=10em,height=4.2em]{\textbf{Model}}{\textbf{Rew. \%$\downarrow$ (Ori.)}}
    } 
&\multicolumn{4}{c}{\textbf{Readability}} & \multicolumn{8}{c}{\textbf{Grammatical Correctness}} \\
\cmidrule(lr){2-5}
\cmidrule(lr){6-13}
& \multirow{2}{*}{\textbf{PopQA}} &\multirow{2}{*}{\textbf{NQ}} & \multirow{2}{*}{\textbf{MARCO}} & \multirow{2}{*}{\textbf{Entity}} & \multicolumn{2}{c}{\textbf{PopQA}} & \multicolumn{2}{c}{\textbf{NQ}} & \multicolumn{2}{c}{\textbf{MARCO}} & \multicolumn{2}{c}{\textbf{Entity}} \\
\cmidrule(lr){6-7}
\cmidrule(lr){8-9}
\cmidrule(lr){10-11}
\cmidrule(lr){12-13}
& & & & & RTT & Typos & RTT & Typos & RTT & Typos & RTT & Typos \\

\midrule

gemma-2-2b-it           & \textbf{20.71} \tiny{(\textcolor{gray}{0.52})} & 16.82 \tiny{(\textcolor{gray}{0.45})} &17.44 \tiny{(\textcolor{gray}{0.20})} &16.43 \tiny{(\textcolor{gray}{0.51})} &\textbf{32.73}& \textbf{20.26} \tiny{(\textcolor{gray}{0.49})} & 28.09& 10.44 \tiny{(\textcolor{gray}{0.43})} & 29.81& 10.19 \tiny{(\textcolor{gray}{0.16})} & 22.47& 10.36 \tiny{(\textcolor{gray}{0.54})}\\
gemma-2-9b-it           & \textbf{17.12} \tiny{(\textcolor{gray}{0.54})} & 13.95 \tiny{(\textcolor{gray}{0.49})} &11.13 \tiny{(\textcolor{gray}{0.20})} &15.25 \tiny{(\textcolor{gray}{0.54})} &\textbf{33.90}& \textbf{19.60} \tiny{(\textcolor{gray}{0.51})} & 24.50& 8.19 \tiny{(\textcolor{gray}{0.48})} & 25.69& 6.98 \tiny{(\textcolor{gray}{0.16})} & 22.92& 10.39 \tiny{(\textcolor{gray}{0.56})}\\
gemma-2-27b-it          & \textbf{16.15} \tiny{(\textcolor{gray}{0.56})} & 11.33 \tiny{(\textcolor{gray}{0.52})} &8.33 \tiny{(\textcolor{gray}{0.20})} &12.68 \tiny{(\textcolor{gray}{0.55})} &\textbf{32.81}& \textbf{18.21} \tiny{(\textcolor{gray}{0.53})} & 24.03& 6.81 \tiny{(\textcolor{gray}{0.51})} & 27.71& 3.79 \tiny{(\textcolor{gray}{0.15})} & 20.73& 10.69 \tiny{(\textcolor{gray}{0.58})}\\
Llama-3.1-8B-Instruct   & \textbf{17.27} \tiny{(\textcolor{gray}{0.55})} & 15.07 \tiny{(\textcolor{gray}{0.52})} &15.23 \tiny{(\textcolor{gray}{0.22})} &13.30 \tiny{(\textcolor{gray}{0.52})} &\textbf{35.27} & \textbf{21.95} \tiny{(\textcolor{gray}{0.50})} & 27.45& 10.54 \tiny{(\textcolor{gray}{0.50})} & 32.37& 12.37 \tiny{(\textcolor{gray}{0.19})} & 25.06& 12.53 \tiny{(\textcolor{gray}{0.56})}\\
Llama-3.1-70B-Instruct  & \textbf{17.66} \tiny{(\textcolor{gray}{0.58})} & 15.77 \tiny{(\textcolor{gray}{0.54})} &13.30 \tiny{(\textcolor{gray}{0.22})} &16.09 \tiny{(\textcolor{gray}{0.55})} &\textbf{35.30}& \textbf{19.49} \tiny{(\textcolor{gray}{0.54})} & 25.43& 8.80 \tiny{(\textcolor{gray}{0.52})} & 24.83& 5.38 \tiny{(\textcolor{gray}{0.17})} & 21.61& 12.71 \tiny{(\textcolor{gray}{0.57})}\\
Qwen2.5-3B-Instruct     & 22.35 \tiny{(\textcolor{gray}{0.47})} & 18.65 \tiny{(\textcolor{gray}{0.42})} &13.27 \tiny{(\textcolor{gray}{0.21})} &\textbf{24.14} \tiny{(\textcolor{gray}{0.46})} &\textbf{32.89}& \textbf{23.77} \tiny{(\textcolor{gray}{0.44})} & 30.74& 15.65 \tiny{(\textcolor{gray}{0.44})} & 32.14& 12.31 \tiny{(\textcolor{gray}{0.17})} & 24.53& 13.20 \tiny{(\textcolor{gray}{0.49})}\\
Qwen2.5-7B-Instruct     & 19.50 \tiny{(\textcolor{gray}{0.53})} & 17.15 \tiny{(\textcolor{gray}{0.48})} &11.55 \tiny{(\textcolor{gray}{0.21})} &\textbf{20.35} \tiny{(\textcolor{gray}{0.51})} &\textbf{34.06}& \textbf{22.67} \tiny{(\textcolor{gray}{0.49})} & 29.67& 13.19 \tiny{(\textcolor{gray}{0.49})} & 24.28& 6.64 \tiny{(\textcolor{gray}{0.17})} & 21.77& 11.99 \tiny{(\textcolor{gray}{0.55})}\\
Qwen2.5-32B-Instruct    & 17.00 \tiny{(\textcolor{gray}{0.56})} & 14.01 \tiny{(\textcolor{gray}{0.52})} &\textbf{32.10} \tiny{(\textcolor{gray}{0.28})} &18.41 \tiny{(\textcolor{gray}{0.53})} &\textbf{34.11}& \textbf{19.69} \tiny{(\textcolor{gray}{0.52})} & 24.80& 9.28 \tiny{(\textcolor{gray}{0.52})} & 26.67& 7.31 \tiny{(\textcolor{gray}{0.16})} & 20.96& 11.62 \tiny{(\textcolor{gray}{0.57})}\\
Qwen2.5-72B-Instruct    & 16.25 \tiny{(\textcolor{gray}{0.56})} & 13.62 \tiny{(\textcolor{gray}{0.54})} &8.36 \tiny{(\textcolor{gray}{0.20})} &\textbf{17.03} \tiny{(\textcolor{gray}{0.54})} &\textbf{33.68}& \textbf{19.27} \tiny{(\textcolor{gray}{0.53})} & 24.28& 7.74 \tiny{(\textcolor{gray}{0.53})} & 27.47& 8.22 \tiny{(\textcolor{gray}{0.16})} & 19.56& 10.35 \tiny{(\textcolor{gray}{0.58})}\\

\midrule
Avg                     & \textbf{18.22} \tiny{(\textcolor{gray}{0.54})} & 15.15 \tiny{(\textcolor{gray}{0.50})} &14.52 \tiny{(\textcolor{gray}{0.22})} &17.07 \tiny{(\textcolor{gray}{0.52})} &\textbf{33.86}& \textbf{20.55} \tiny{(\textcolor{gray}{0.51})} & 26.55& 10.07 \tiny{(\textcolor{gray}{0.49})} & 27.88& 8.13 \tiny{(\textcolor{gray}{0.17})} & 22.18& 11.54 \tiny{(\textcolor{gray}{0.56})}\\

\bottomrule
\end{tabular}
}
\end{subtable}

\vspace{1em}

\begin{subtable}{\textwidth}
\centering
\begin{adjustbox}{width=0.85\textwidth}
\begin{tabular}{lcccccccc}
\toprule
 \multirow{2}{*}{%
        \diagbox[width=10em,height=2.8em]{\textbf{Model}}{\textbf{Rew. \%$\downarrow$ (Ori.)}}
    } 
 & \multicolumn{4}{c}{\textbf{Formality}} & \multicolumn{4}{c}{\textbf{Politeness}} \\

\cmidrule(lr){2-5} \cmidrule(lr){6-9}

& \textbf{PopQA} & \textbf{NQ} & \textbf{MARCO} & \textbf{Entity} & \textbf{PopQA} & \textbf{NQ} & \textbf{MARCO} & \textbf{Entity} \\
\midrule

gemma-2-2b-it           & 13.64 \tiny{(\textcolor{gray}{0.61})} &  12.44 \tiny{(\textcolor{gray}{0.46})} & \textbf{23.31} \tiny{(\textcolor{gray}{0.17})} & 7.66 \tiny{(\textcolor{gray}{0.60})} & 8.37 \tiny{(\textcolor{gray}{0.52})} & 4.42 \tiny{(\textcolor{gray}{0.44})} & \textbf{11.16} \tiny{(\textcolor{gray}{0.20})} & 2.96 \tiny{(\textcolor{gray}{0.59})} \\
gemma-2-9b-it           & 11.67 \tiny{(\textcolor{gray}{0.64})} &  9.95  \tiny{(\textcolor{gray}{0.50})} & \textbf{19.68} \tiny{(\textcolor{gray}{0.17})} & 8.85 \tiny{(\textcolor{gray}{0.62})} & \textbf{8.54} \tiny{(\textcolor{gray}{0.54})} & 3.60 \tiny{(\textcolor{gray}{0.48})} & 8.19 \tiny{(\textcolor{gray}{0.20})} & 3.01 \tiny{(\textcolor{gray}{0.61})} \\
gemma-2-27b-it          & 12.21 \tiny{(\textcolor{gray}{0.65})} &  9.92  \tiny{(\textcolor{gray}{0.52})} & \textbf{19.13} \tiny{(\textcolor{gray}{0.17})} & 7.19 \tiny{(\textcolor{gray}{0.63})} & \textbf{7.93} \tiny{(\textcolor{gray}{0.56})} & 3.38 \tiny{(\textcolor{gray}{0.51})} & 6.13 \tiny{(\textcolor{gray}{0.20})} & 1.97 \tiny{(\textcolor{gray}{0.61})} \\
Llama-3.1-8B-Instruct   & 12.23 \tiny{(\textcolor{gray}{0.65})} &  10.51 \tiny{(\textcolor{gray}{0.52})} & \textbf{17.59} \tiny{(\textcolor{gray}{0.20})} & 7.16 \tiny{(\textcolor{gray}{0.62})} & 7.59 \tiny{(\textcolor{gray}{0.55})} & 3.46 \tiny{(\textcolor{gray}{0.51})} & \textbf{9.90} \tiny{(\textcolor{gray}{0.23})} & 3.50 \tiny{(\textcolor{gray}{0.59})} \\
Llama-3.1-70B-Instruct  & 11.48 \tiny{(\textcolor{gray}{0.67})} &  12.09 \tiny{(\textcolor{gray}{0.54})} & \textbf{19.00} \tiny{(\textcolor{gray}{0.20})} & 7.19 \tiny{(\textcolor{gray}{0.64})} & 7.00 \tiny{(\textcolor{gray}{0.57})} & 6.29 \tiny{(\textcolor{gray}{0.53})} & \textbf{9.89} \tiny{(\textcolor{gray}{0.23})} & 3.61 \tiny{(\textcolor{gray}{0.61})} \\
Qwen2.5-3B-Instruct     & 10.97 \tiny{(\textcolor{gray}{0.58})} &  11.38 \tiny{(\textcolor{gray}{0.45})} & \textbf{20.87} \tiny{(\textcolor{gray}{0.18})} & 7.51 \tiny{(\textcolor{gray}{0.56})} & 11.54 \tiny{(\textcolor{gray}{0.47})} & 5.91 \tiny{(\textcolor{gray}{0.42})} & \textbf{12.30} \tiny{(\textcolor{gray}{0.21})} & 7.74 \tiny{(\textcolor{gray}{0.55})} \\
Qwen2.5-7B-Instruct     & 12.99 \tiny{(\textcolor{gray}{0.63})} &  11.46 \tiny{(\textcolor{gray}{0.50})} & \textbf{19.84} \tiny{(\textcolor{gray}{0.18})} & 8.08 \tiny{(\textcolor{gray}{0.61})} & \textbf{10.76} \tiny{(\textcolor{gray}{0.54})} & 5.05 \tiny{(\textcolor{gray}{0.48})} & 7.08 \tiny{(\textcolor{gray}{0.21})} & 4.69 \tiny{(\textcolor{gray}{0.60})} \\
Qwen2.5-32B-Instruct    & 11.34 \tiny{(\textcolor{gray}{0.65})} &  7.85  \tiny{(\textcolor{gray}{0.53})} & \textbf{14.80} \tiny{(\textcolor{gray}{0.18})} & 6.24 \tiny{(\textcolor{gray}{0.62})} & \textbf{8.65} \tiny{(\textcolor{gray}{0.56})} & 4.76 \tiny{(\textcolor{gray}{0.52})} & 6.66 \tiny{(\textcolor{gray}{0.21})} & 5.64 \tiny{(\textcolor{gray}{0.60})} \\
Qwen2.5-72B-Instruct    & 10.68 \tiny{(\textcolor{gray}{0.66})} &  9.11  \tiny{(\textcolor{gray}{0.55})} & \textbf{20.72} \tiny{(\textcolor{gray}{0.18})} & 5.75 \tiny{(\textcolor{gray}{0.63})} & 7.36 \tiny{(\textcolor{gray}{0.56})} & 5.45 \tiny{(\textcolor{gray}{0.54})} & \textbf{16.62} \tiny{(\textcolor{gray}{0.20})} & 3.86 \tiny{(\textcolor{gray}{0.62})} \\

\midrule
Avg    & 11.91 \tiny{(\textcolor{gray}{0.64})} &  10.52  \tiny{(\textcolor{gray}{0.51})} & \textbf{19.44} \tiny{(\textcolor{gray}{0.18})} & 7.29 \tiny{(\textcolor{gray}{0.61})} & 8.64 \tiny{(\textcolor{gray}{0.54})} & 4.70 \tiny{(\textcolor{gray}{0.49})} & \textbf{9.77} \tiny{(\textcolor{gray}{0.21})} & 4.11 \tiny{(\textcolor{gray}{0.60})} \\

\bottomrule
\end{tabular}\end{adjustbox}
\end{subtable}
\vspace{1em}

\caption{RAG performance on answer match (AM) scores using ModernBERT retriever with the Gemma 2, Llama 3.1, and Qwen 2.5 model families across four datasets.
Results show relative percentage performance degradation on rewritten queries (Rew. \% $\downarrow$) and \textcolor{gray}{the original query performance (Ori.)} within parentheses in gray.
For RTT, it has the same original scores as Typos.
\textbf{The largest degradation value} among four datasets is in bold.
\textbf{All systems exhibit performance drops across all linguistic variations and datasets.}}
\label{tab:generation}
\end{table*}

\paragraph{Grammatical variations have the highest impact on retrieval performance.}
On average, grammaticality rewrites emerge as the most impactful linguistic variation on recall performance. Round-trip translation degrades recall by an average of 19.56\% across all datasets. Interestingly, ModernBERT shows greater vulnerability to these structural transformations (20.12\% drop) compared to Contriever (19\% drop). Typographical errors present another significant challenge, causing an average recall reduction of 18.51\%. However, the retrievers display opposite behavior patterns with typos: Contriever exhibits substantially lower robustness (22.22\% drop) than ModernBERT (14.81\% drop).  
This suggests that ModernBERT's diverse training data mixture likely enables it to develop greater robustness to character-level grammatical perturbations compared to Contriever.

\paragraph{Politeness variations have minimal impact on retrieval performance.}
Politeness variations have the least impact on retrieval performance, with an average recall drop of only 6.48\% across all datasets and retrievers. This stands in stark contrast to grammatical variations (19.56\%) and typos (18.51\%). The minimal effect is most evident in Natural Questions with Contriever (1.16\%) and EntityQuestions with Contriever (1.70\%). This suggests that retrieval models effectively filter out social courtesy markers while preserving their focus on the query's core semantic content and keywords, maintaining robust performance despite changes in query politeness level.

\paragraph{Retrieval performance drops independent of query length.}
Our analysis, as shown in Table \ref{tab:retriever_performance} demonstrates that query length changes do not directly correlate with retrieval performance. Queries with increased formality showed substantial length increases (+13.65 tokens) yet produced inconsistent performance impacts across datasets. Conversely, round-trip translated queries were marginally shorter (-0.26 tokens) but consistently caused significant performance degradation. This indicates retrieval models respond more to linguistic quality (grammatical correctness, readability) than to query length itself, highlighting the need for systems robust to linguistic variations rather than optimized for specific query lengths. We confirm this by analyzing semantic preservation using LLM-as-a-judge in Table \ref{tab:llm-as-a-judge} and a human evaluation of results in Table \ref{tab:semantic_preservation_qual}.

\paragraph{Scaling up number of documents improves performance.}
Table~\ref{tab:retriever_performance} shows performance degradation ($\Delta$R@K) decreasing as $K$ increases, indicating linguistic perturbations cause relevant documents to slide down rather than disappear from the ranked list. As an example, for rewrites with typos, $\Delta$ R@K for Contriever decreases from 22.2 at R@5 to 12.0 at R@100, and for ModernBERT from 20.1 to 9.8. This is detailed further in Appendix \ref{sec:scaling}. This ranking deterioration forces downstream language models to operate with suboptimal information, potentially compromising response quality. We further investigate this hypothesis in Section \ref{sec:reranker} by examining if rerankers can improve recall scores for Top-5 retrieved documents.

\begin{table*}[t]
\centering
\small
\begin{tabular}{lccccc}
\toprule
\textbf{Dataset} & \textbf{Readability} & \textbf{Formality} & \textbf{Politeness} & \textbf{Gram-RTT} & \textbf{Gram-Typos} \\
\midrule
PopQA              & 19.41\% & 11.79\% & 6.75\% & 35.60\% & 21.20\% \\
NQ & 13.12\% & 4.27\%  & 4.38\% & 22.74\% & 5.63\%  \\
MARCO           & 10.56\% & 4.05\%  & 1.75\% & 15.21\% & 5.32\%  \\
Entity    & 13.20\% & 5.06\%  & 2.86\% & 22.18\% & 10.49\% \\
\bottomrule
\end{tabular}
\caption{RAG performance in LLM-as-a-Judge evaluation (ModernBERT + Qwen2.5-72B-Instruct), with relative percentage performance degradation.
\textbf{LLM-as-A-Judge evaluation shows a consistent degradation pattern.}}
\label{tab:llm-as-a-judge}
\end{table*}

\subsection{Generation Analysis}

The RAG experiment results on ModernBERT retriever and nine LLMs are presented in Table~\ref{tab:generation} with answer match (AM) scores. 
Table~\ref{tab:llm-as-a-judge} shows the results with LLM-as-a-Judge evaluation.
Overall, RAG systems show performance degradation on all linguistic variations.
\looseness=-1

\paragraph{RAG systems are sensitive to linguistic variations.}
As illustrated in Table~\ref{tab:generation}, across all datasets, we observe a noticeable overall degradation in performance when queries are rewritten to become less formal, more polite, less readable, or have grammatical errors.
Across all datasets, linguistic dimensions, and experimental settings, we found average drops of 16.52\% (AM), 41.15\% (EM), and 19.60\% (F1).
The PopQA dataset shows the highest sensitivity to all linguistic variations, with an average performance drop of 18.64\% on AM scores.
Particularly notable were the effects of reduced readability (18.22\% degradation) and round-trip translation (33.86\% degradation).
These findings suggest that while the RAG systems perform well on standard NLP benchmarks with structured queries, they remain vulnerable to common linguistic variations.
The full experiment results are shown in Appendix~\ref{app:full_rag_results}.

\paragraph{Politeness reformulations yield different impacts on AM and EM scores.}
When queries are rephrased to be more polite, we find that the AM scores remain relatively similar to those of the original, with less than 10\% change for all datasets. However, there are significant drops in exact match (EM) scores. 
Specifically, we observe 44.57\% and 18.32\% drops in EM scores for queries from the Natural Questions and PopQA datasets, respectively. 
Although the rewritten queries preserve the original pragmatic intent, they introduce longer phrasing and auxiliary words that shift embeddings and sometimes nudge generators toward more verbose outputs. As a result, they still cause slight regressions in the RAG.
% Through manual checks, we find that LLMs tend to generate more complete or formal responses in polite query formulations, which results in lower EM rates but similar performance on AM.

\paragraph{The round-trip translation errors and typos highlight different sensitivities.}
Round-trip translation, which introduces structural sentence transformations, generally causes notable decreases across all datasets, showing 33.86\%, 26.55\%, 27.88\%, and 22.18\% drops in AM scores in the PopQA, Natural Questions, MS MARCO, and EntityQuestions, respectively. 
In contrast, typos, mainly introducing surface-level grammatical errors, produce moderate but less drastic performance degradation. 
This finding is consistent with the retrieval experiment results, suggesting that RAG systems are more vulnerable to structural transformations than superficial grammatical mistakes.
\looseness=-1

\subsection{Retrieval Method and LLM Scale Influence}

\begin{figure}[hbt]
    \centering
    \includegraphics[width=\linewidth]{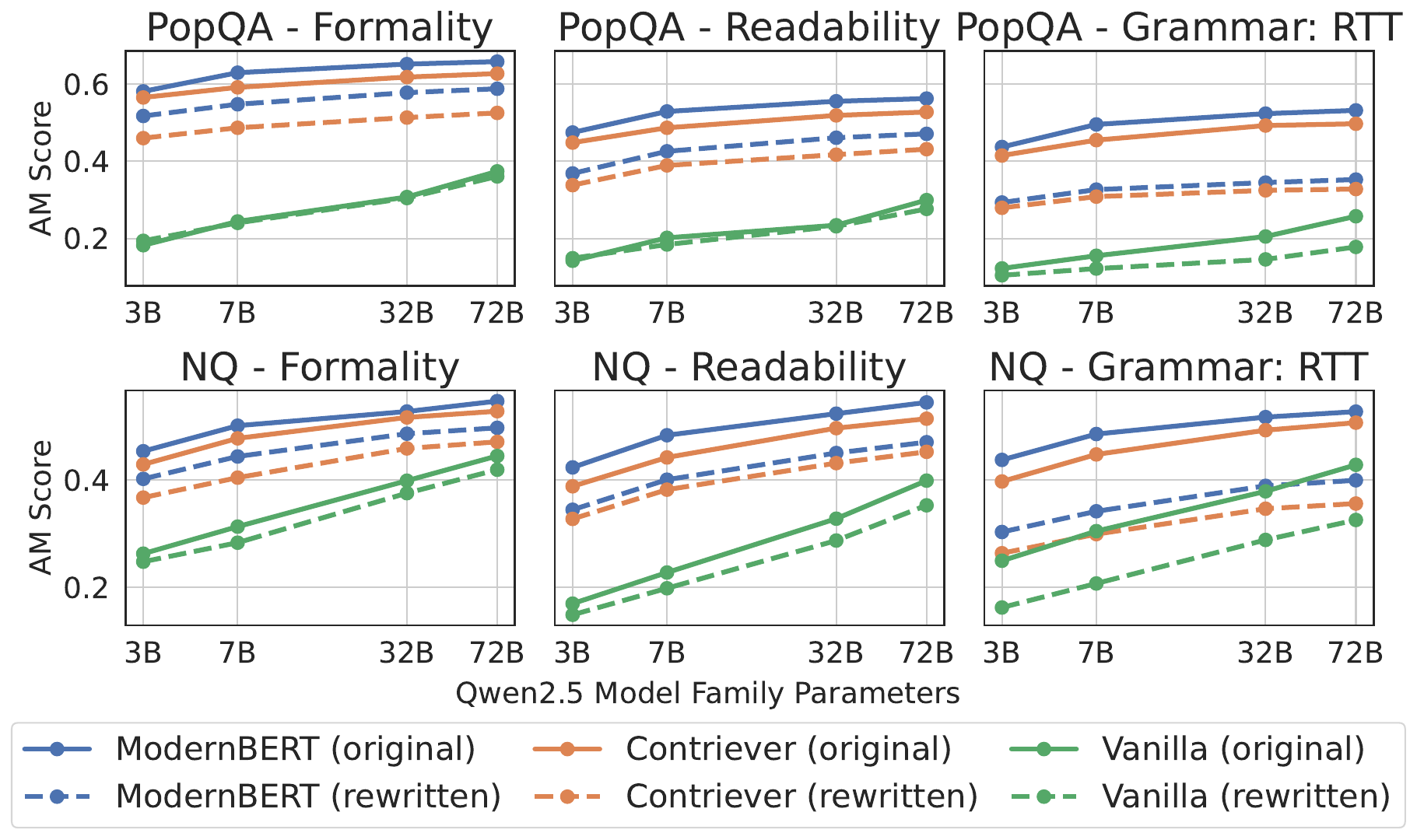}
    \caption{PopQA and Natural Questions (NQ) LLMs scaling results, augmented with ModernBERT, Contriever, and LLM-only generation (Vanilla). 
    \textbf{Retrieval-augmented generation is more sensitive to linguistic variations than the LLM-only generation.}}
    \label{fig:retr_comp_gen}
\end{figure}

In this section, we are going to investigate the influence of different retrieval methods and LLM scales on the robustness of the RAG systems. The main results are shown in Figure~\ref{fig:retr_comp_gen}.

\paragraph{RAG systems with ModernBERT retrieval show greater robustness to linguistic variations.}
As shown in Figure~\ref{fig:retr_comp_gen}, generation results based on ModernBERT retrieval consistently outperform those with Contriever retrieval across both original and rewritten queries.
Notably, RAG systems with ModernBERT demonstrate superior robustness to linguistic variations, exhibiting an average performance drop of only 19.52\% on rewritten queries compared to 24.38\% for Contriever. 
This suggests ModernBERT retrieval maintains better semantic understanding when handling linguistically varied inputs.

\paragraph{RAG systems show higher sensitivity to linguistic variations than LLM-only generations.}
For PopQA, we observe an average performance drop of 22.53\% across all linguistic variations and both retrieval models, while LLM-only generations experience only a 10.78\% reduction. 
Even more striking, Figure~\ref{fig:retr_comp_gen} shows that there is barely any performance difference in LLM-only generation on formality rewrites, which suggests that errors are cascaded from the retrieval component to the generation component in the RAG system.
These findings further indicate that retrieval components represent the primary vulnerability in RAG systems when handling linguistic variations.

\paragraph{LLM scaling doesn't always help with mitigating performance gaps in RAG systems.}
Notably, the performance gap between original and rewritten queries narrows for formality and readability variations as LLMs scale up (see Figure~\ref{fig:retr_comp_gen}).
Specifically, PopQA shows reduced degradation on less readable queries from 22.35\% at 3B to 16.25\% at 72B parameter.
This suggests that larger models can extract relevant information better from retrieved contexts.
However, this scaling benefit remains selective and limited; for round-trip translation variations, the performance gap actually widens with increased model size. 
This counterintuitive finding may be attributed to the structural transformations introduced during translation that become more problematic for larger models attempting more precise reasoning.

\paragraph{Human annotator evaluation aligns with our automated evaluation metrics}
To validate our automated metrics and control for potential confounds such as length or verbosity, two independent annotators evaluated 250 samples across EntityQuestions (100 samples) and NaturalQuestions (150 samples). For typos and RTT variations, rewrites with more than 2 character edits to any entity were marked as not preserving semantic meaning. As shown in Table~\ref{tab:semantic_preservation_qual}, semantic preservation rates ranged from 86.7\% to 100\% with high inter-annotator agreement (96.7\%-100\%), confirming that observed performance differences stem from linguistic variation rather than semantic drift. Full annotation protocols are provided in Section~\ref{sec:semantic_preservation}.

\subsection{Exploring the Robustness of Advanced RAG Systems}

Many modern RAG systems include more components than simply retrieval and generation, which aim to make them more useful for users \citep{gao2022precisezeroshotdenseretrieval}. 
In this section, we explore 
the possibility of using a simple \textbf{query-expansion} step to fix the vulnerability on linguistic variations and
whether the addition of \textbf{reranking} improves RAG robustness. 
Detailed results of the experiment can be found in Appendices \ref{app:hyde} and \ref{app:rerank}.

\begin{figure}
    \centering
    \includegraphics[width=\linewidth]{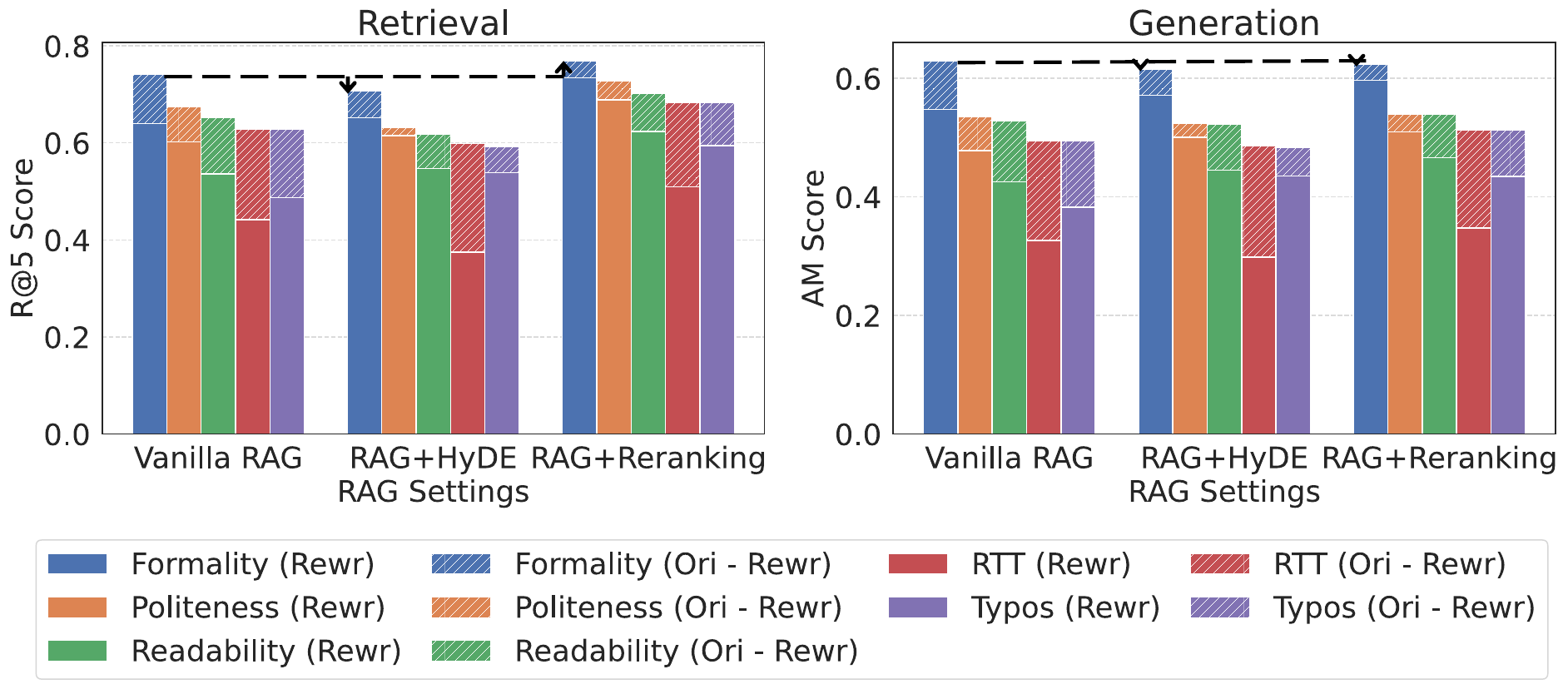}
    \caption{Retrieval (ModernBERT, R@5 Score) and generation (Qwen2.5-7B-Instruct, AM Score) performance across different RAG settings on PopQA. \textbf{We find that (1) adding HyDE and Rerank to the RAG pipeline improves the robustness to linguistic variations, but still lags behind original queries in performance. (2) HyDE improves robustness but slightly reduces performance on original queries. (3) Reranking improves performance on both original and rewritten queries.}  }
   
    \label{fig:hyde-rerank}
\end{figure}

\subsubsection{Query Expansion}
\label{sec:rerank}

We evaluate Hypothetical Document Embeddings (HyDE; \citealt{gao2022precisezeroshotdenseretrieval}) for query expansion. Figure~\ref{fig:hyde-rerank} reveals that HyDE improves ModernBERT's retrieval on linguistically varied queries (readability, typos, formality, politeness) by 2.61\% on average, but severely impairs performance on round-trip translated queries (11\% decrease). For original queries, HyDE consistently reduces ModernBERT's retrieval effectiveness by 5.43\%. Similarly, generation quality increases for rewritten queries across PopQA (3.77\%) but decreases for original queries (1.92\%). These findings suggest HyDE provides insufficient benefits for ModernBERT, underscoring the need for more effective query expansion methods.

\subsubsection{Reranker}
\label{sec:reranker}

The retriever must be efficient for large document collections containing millions of entries, although it may sometimes retrieve irrelevant candidates. 
To address this, we incorporate a Cross-Encoder-based re-ranker to significantly enhance the quality of final answers. 
Specifically, we employ the MS MARCO Cross-Encoders developed by \citet{reimers-2019-sentence-bert} to re-rank passages retrieved by ModernBERT and Contriever. 
As illustrated in Figure~\ref{fig:hyde-rerank}, re-ranking substantially improves retrieval performance, particularly for rewritten queries, achieving an average improvement of 16.56\% compared to only 7.40\% for original queries. 
In contrast, generation results show a modest improvement of 1.83\% for original queries and a more substantial improvement of 8.50\% for rewritten queries. 
These findings suggest that the effectiveness of re-ranking is especially pronounced when handling rewritten queries and highlight the importance of improving the robustness of retrieval systems.

%% file: sections/06_Conclusion.tex
\section{Conclusion}
We conduct the first large-scale, systematic investigation into how linguistic variations—specifically formality, readability, politeness, and grammatical correctness—impact the robustness of RAG systems. 
Our analysis reveals that both the retrieval and generation components suffer performance degradation when faced with linguistic variations.
Notably, RAG systems exhibit greater vulnerability to linguistic variations compared to LLM-only generations, indicating potential cascading errors within the retrieval-generation pipeline. 
Crucially, increasing the scale of LLMs does not consistently mitigate these robustness issues, and even advanced retrieval techniques like HyDE and reranking show similar susceptibility. 
These findings highlight the need to develop strategies that ensure reliable performance across linguistically varied queries, guiding future improvements of real-world RAG systems.

%% file: sections/07_Appendix.tex
\section{Linguistic Rewriting Settings}
\subsection{Linguistic Variations}
\label{sec:app_ling}

In this section, we detail our design of linguistic variations as well as our qualitative analysis of the queries we generated.

\paragraph{Query rewrite design}: For each linguistic dimension, we establish quantitative thresholds to ensure meaningful rewrites from original query $x_i$ to the rewritten query $x_i'$:

\begin{itemize}
    \item Formality: $x_i'$ must score below 0.5 probability using our formality classifier
    \item Readability: $x_i'$ must score below 60 on the Flesch Reading Ease Score
    \item Politeness: The sum of \textit{somewhat polite} and \textit{polite} score on $x_i'$ must above 0.5 using Polite Guard classification model.
    \item Grammatical correctness (Typos): $x_i'$ must have edit distance $\geq 1$ from $x_i$ and must not have the same GLEU score. Edit distance between two strings is the minimum number of operations required to transform one string into the other.
    \item Grammatical correctness (Round-trip translation): $x_i'$ must differ from $x_i$ after English-Afrikaans-English translation, constrained by ensuring that both queries do not have the same GLEU score and an edit distance of 0.
\end{itemize}

Each rewrite for each prompt must adhere to these thresholds in order to develop the final dataset we use. 
In addition to this, we want to ensure that the rewritten queries are semantically similar to the original query. Therefore, in the query-rewriting process, we add an additional constraint where the sentence similarity between the original and rewritten queries must be greater than 0.7, as assessed by the MPNet-v2 model \cite{song2020mpnetmaskedpermutedpretraining} in the SentenceBert Library \cite{reimers-2019-sentence-bert}.

\paragraph{Evaluation}: Flesch Reading score is defined as follows:

% \begin{equation}
% \text{FRES} = 206.835 - 1.015 \left(\frac{\text{total words}}{\text{total sentences}}\right)
% - 84.6 \left(\frac{\text{total syllables}}{\text{total words}}\right)
% \end{equation}

\begin{equation}
\begin{split}
\mathrm{FRES}
&= 206.835
  - 1.015\;\frac{\text{total words}}{\text{total sentences}}\\
&\quad
  - 84.6\;\frac{\text{total syllables}}{\text{total words}}
\end{split}
\end{equation}

\subsection{Semantic Preservation of Linguistic Variations}
\label{sec:semantic_preservation}
To ensure our linguistic rewrites preserve semantic meaning, we conducted a qualitative evaluation where two independent researchers annotated randomly sampled query rewrites per linguistic property from EntityQuestions (100 examples) and Natural Questions (150 examples) datasets, totaling 250 queries. 

\subsubsection{Annotation Guidelines}
Annotators were provided with the following protocol to assess whether each rewritten query preserved the semantic meaning of the original:

\begin{itemize}
    \item Mark as \textit{semantically preserved} if both queries would elicit identical factual answers
    \item Mark as \textit{not preserved} if the rewrite changes entities, relationships, or temporal/spatial constraints
    \item Rule for Typos and RTT: If more than 2 characters are edited in any entity mention, mark as \textit{not preserved}. This threshold helps maintain a balance between introducing realistic linguistic variations and preserving essential meaning.
    \item Ignore stylistic differences that do not affect the core information need
\end{itemize}

Annotators worked independently and their judgments were compared to compute inter-annotator agreement.

% \begin{table}[t]
% \centering
% \small
% \setlength{\tabcolsep}{6pt}
% \begin{subtable}{\columnwidth}
% \centering
% \begin{tabular}{lcc}
% \hline
% \textbf{Linguistic Variation} & \textbf{Semantic } & \textbf{Inter-annotator} \\
% & \textbf{Preservation (\%)} & \textbf{ Agreement (\%)} \\
% \hline
% Politeness  & 100.0 & 100.0 \\
% Formality   & 98.6  & 100.0 \\
% Readability & 96.7  & 96.7  \\
% Typos       & 95.0  & 100.0 \\
% RTT         & 95.0  & 100.0 \\
% \hline
% \end{tabular}
% \caption{EntityQuestions}
% \end{subtable}
% \vspace{0.8em}
% \begin{subtable}{\linewidth}
% \centering
% \begin{tabular}{lcc}
% \hline
% \textbf{Linguistic Variation} & \textbf{Semantic } & \textbf{Inter-annotator} \\
% & \textbf{Preservation (\%)} & \textbf{ Agreement (\%)} \\
% \hline
% Politeness  & 100.0 & 100.0 \\
% Formality   & 90.0  & 96.7  \\
% Readability & 100.0 & 100.0 \\
% Typos       & 86.7  & 96.7  \\
% RTT         & 96.7  & 96.7  \\
% \hline
% \end{tabular}
% \caption{Natural Questions}
% \end{subtable}
% \caption{Percentage of queries that preserve semantic meaning after linguistic rewrites, along with inter-annotator agreement rates across 250 annotated samples.}
% \label{tab:semantic_preservation_qual}
% \end{table}
\begin{table}[t]
\centering
\scriptsize
\setlength{\tabcolsep}{3pt}
\renewcommand{\arraystretch}{1.05}

\begin{subtable}{\columnwidth}
\centering
\resizebox{\columnwidth}{!}{%
\begin{tabular}{lcc}
\hline
\textbf{Linguistic Variation} & \textbf{Semantic} & \textbf{Inter-annotator} \\
                             & \textbf{Preservation (\%)} & \textbf{Agreement (\%)} \\
\hline
Politeness  & 100.0 & 100.0 \\
Formality   & 98.6  & 100.0 \\
Readability & 96.7  & 96.7  \\
Typos       & 95.0  & 100.0 \\
RTT         & 95.0  & 100.0 \\
\hline
\end{tabular}%
}
\caption{EntityQuestions}
\end{subtable}

\vspace{0.6em}

\begin{subtable}{\columnwidth}
\centering
\resizebox{\columnwidth}{!}{%
\begin{tabular}{lcc}
\hline
\textbf{Linguistic Variation} & \textbf{Semantic} & \textbf{Inter-annotator} \\
                             & \textbf{Preservation (\%)} & \textbf{Agreement (\%)} \\
\hline
Politeness  & 100.0 & 100.0 \\
Formality   & 90.0  & 96.7  \\
Readability & 100.0 & 100.0 \\
Typos       & 86.7  & 96.7  \\
RTT         & 96.7  & 96.7  \\
\hline
\end{tabular}%
}
\caption{Natural Questions}
\end{subtable}

\caption{Percentage of queries that preserve semantic meaning after linguistic rewrites, along with inter-annotator agreement rates across 250 annotated samples.}
\label{tab:semantic_preservation_qual}
\end{table}

\subsubsection{Results and Analysis}
Table~\ref{tab:semantic_preservation_qual} shows high semantic preservation rates across all linguistic dimensions, ranging from 86.7\% to 100\%. The high inter-annotator agreement (96.7\%-100\%) further validates the reliability of our assessment approach. These results demonstrate the robustness of our rewriting methodology in preserving semantic content while introducing natural linguistic variations. Notably, even with our strict 2-character edit threshold for typos and RTT, we achieve high preservation rates (86.7\%-96.7\%), indicating that our generation process successfully balances realism with semantic fidelity. This semantic preservation is critical for our experimental design, as it ensures that performance differences observed in RAG systems can be attributed to linguistic variations rather than semantic drift.

\section{Detailed RAG Experimental Settings}

\subsection{LLMs Generation Hyperparameter Setting}\label{app:gen_setting}
We set \texttt{temperature} as 0.5 and \texttt{top\_q} 0.90 to strike a balance between creativity and accuracy. The \texttt{max\_tokens} is set to 128 considering the gold answers length.

\section{Experiment Results on Llama 3.1 Rewriting Queries}\label{app:llama_rewriting}

We conducted supplementary experiments by sampling 500 queries from the PopQA dataset. We generated rewritten queries using Llama-3.1-70B-Instruct, employing the same rewriting criteria as previously described. Then, we performed RAG experiments using ModernBERT as the retriever and Qwen2.5-7B-Instruct as the generator. The results are summarized below:

\begin{table}[h!]
\centering
\begin{tabular}{lccc}
\toprule
\textbf{Dimension} & \textbf{Original} & \textbf{Rewritten} & $\boldsymbol{\Delta}$ \\
\midrule
Readability & 0.624 & 0.490 & \textbf{21.5\%} \\
Formality   & 0.826 & 0.746 & 9.7\% \\
Politeness  & 0.728 & 0.626 & 14.0\% \\
\bottomrule
\end{tabular}
\caption{Retrieval Performance (R@5)}
\end{table}

\begin{table}[h!]
\centering
\begin{tabular}{lccc}
\toprule
\textbf{Dimension} & \textbf{Original} & \textbf{Rewritten} & $\boldsymbol{\Delta}$ \\
\midrule
Readability & 0.500 & 0.372 & \textbf{25.6\%} \\
Formality   & 0.776 & 0.684 & 11.9\% \\
Politeness  & 0.604 & 0.520 & 13.9\% \\
\bottomrule
\end{tabular}
\caption{RAG Generation Performance (Answer Match - AM Score)}
\end{table}

In general, \textbf{the RAG system remains notably sensitive to the linguistic variations introduced by LLaMA}. Specifically, queries rewritten for reduced readability caused the most significant performance degradation, with 21.5\% in retrieval accuracy and 25.6\% in the AM score. Unlike variations introduced by GPT, the system exhibited greater sensitivity to changes in politeness compared to formality. We will present a comprehensive analysis and more detailed results in the revised paper.

\section{Data and Code Availability}
Our code and rewritten query datasets will be released after the peer review stage under the CC BY 4.0 license.
All existing datasets, models, and codes used in this work were employed consistently with their intended research purposes.

\section{Full Retrieval Experiment Results}

 In this section we provide the absolute results from Contriever retriever experiments that led to Table \ref{tab:retriever_performance}
\begin{table*}[ht]
\centering
\small
\begin{tabular}{l l r r r r}
\hline
\textbf{Dataset} & \textbf{Linguistics} & \textbf{R@5} & \textbf{R@10} & \textbf{R@20} & \textbf{R@100} \\
\hline
EntityQuestions & RTT                & 0.6838 & 0.7332 & 0.7710 & 0.8422 \\
EntityQuestions & Typos              & 0.6838 & 0.7332 & 0.7710 & 0.8422 \\
EntityQuestions & Formality          & 0.6846 & 0.7292 & 0.7616 & 0.8240 \\
EntityQuestions & Politeness         & 0.6744 & 0.7218 & 0.7594 & 0.8310 \\
EntityQuestions & Readability        & 0.6434 & 0.6994 & 0.7452 & 0.8354 \\
MS MARCO         & RTT                & 0.3412 & 0.4068 & 0.4694 & 0.6102 \\
MS MARCO         & Typos              & 0.3412 & 0.4068 & 0.4694 & 0.6102 \\
MS MARCO         & Formality          & 0.2534 & 0.3252 & 0.4048 & 0.5598 \\
MS MARCO         & Politeness         & 0.2620 & 0.3414 & 0.4152 & 0.5664 \\
MS MARCO         & Readability        & 0.2512 & 0.3280 & 0.4072 & 0.5718 \\
Natural Questions & RTT               & 0.6246 & 0.7118 & 0.7834 & 0.8822 \\
Natural Questions & Typos             & 0.6246 & 0.7118 & 0.7834 & 0.8822 \\
Natural Questions & Formality         & 0.6472 & 0.7372 & 0.7954 & 0.8868 \\
Natural Questions & Politeness        & 0.6012 & 0.6946 & 0.7588 & 0.8488 \\
Natural Questions & Readability       & 0.5978 & 0.6882 & 0.7582 & 0.8536 \\
PopQA            & RTT               & 0.5938 & 0.6614 & 0.7148 & 0.8192 \\
PopQA            & Typos             & 0.5938 & 0.6614 & 0.7148 & 0.8192 \\
PopQA            & Formality         & 0.6974 & 0.7574 & 0.8066 & 0.8760 \\
PopQA            & Politeness        & 0.6220 & 0.6942 & 0.7576 & 0.8534 \\
PopQA            & Readability       & 0.6108 & 0.6856 & 0.7368 & 0.8344 \\
\hline
\end{tabular}
\caption{Contriever Retrieval performance (R@k) of original queries across datasets and linguistic modifications.}
\end{table*}

\begin{table*}[ht]
\centering
\small
\begin{tabular}{l l r r r r}
\hline
\textbf{Dataset} & \textbf{Linguistics} & \textbf{R@5} & \textbf{R@10} & \textbf{R@20} & \textbf{R@100} \\
\hline
EntityQuestions & RTT                & 0.5842 & 0.6428 & 0.6926 & 0.7896 \\
EntityQuestions & Typos              & 0.5826 & 0.6452 & 0.6964 & 0.7836 \\
EntityQuestions & Formality          & 0.6113 & 0.6626 & 0.7025 & 0.7815 \\
EntityQuestions & Politeness         & 0.6629 & 0.7090 & 0.7507 & 0.8253 \\
EntityQuestions & Readability        & 0.5887 & 0.6471 & 0.6979 & 0.7935 \\
MS MARCO         & RTT                & 0.3100 & 0.3662 & 0.4328 & 0.5754 \\
MS MARCO         & Typos              & 0.2882 & 0.3460 & 0.4052 & 0.5440 \\
MS MARCO         & Formality          & 0.1502 & 0.2020 & 0.2722 & 0.4365 \\
MS MARCO         & Politeness         & 0.2189 & 0.2933 & 0.3764 & 0.5448 \\
MS MARCO         & Readability        & 0.1982 & 0.2716 & 0.3467 & 0.5235 \\
Natural Questions & RTT               & 0.4778 & 0.5718 & 0.6506 & 0.7898 \\
Natural Questions & Typos             & 0.4320 & 0.5230 & 0.6180 & 0.7730 \\
Natural Questions & Formality         & 0.5479 & 0.6440 & 0.7215 & 0.8456 \\
Natural Questions & Politeness        & 0.5942 & 0.6832 & 0.7516 & 0.8443 \\
Natural Questions & Readability       & 0.5519 & 0.6481 & 0.7253 & 0.8346 \\
PopQA            & RTT               & 0.4216 & 0.4854 & 0.5422 & 0.6654 \\
PopQA            & Typos             & 0.4288 & 0.4890 & 0.5482 & 0.6724 \\
PopQA            & Formality         & 0.5582 & 0.6269 & 0.6800 & 0.7856 \\
PopQA            & Politeness        & 0.5704 & 0.6426 & 0.7045 & 0.8081 \\
PopQA            & Readability       & 0.4981 & 0.5635 & 0.6192 & 0.7332 \\
\hline
\end{tabular}
\caption{Contriever Retrieval performance (R@k) of rewritten queries across datasets and linguistic modifications}
\end{table*}
 \clearpage

\begin{table*}[ht]
\centering
\small
\setlength{\tabcolsep}{4pt}
\renewcommand{\arraystretch}{0.85}
\begin{tabular}{l l r r r r}
\toprule
\textbf{Dataset} & \textbf{Linguistics} & \textbf{R@5} & \textbf{R@10} & \textbf{R@20} & \textbf{R@100} \\
\midrule
EntityQuestions & RTT                & 0.6614 & 0.7184 & 0.7598 & 0.8284 \\
EntityQuestions & Typos              & 0.6614 & 0.7184 & 0.7598 & 0.8284 \\
EntityQuestions & Formality          & 0.6798 & 0.7240 & 0.7558 & 0.8150 \\
EntityQuestions & Politeness         & 0.6730 & 0.7214 & 0.7594 & 0.8276 \\
EntityQuestions & Readability        & 0.6132 & 0.6758 & 0.7254 & 0.8108 \\
MS MARCO         & RTT                & 0.3204 & 0.3916 & 0.4574 & 0.5680 \\
MS MARCO         & Typos              & 0.3204 & 0.3916 & 0.4574 & 0.5680 \\
MS MARCO         & Readability        & 0.3982 & 0.4818 & 0.5604 & 0.6746 \\
MS MARCO         & Formality          & 0.4074 & 0.4896 & 0.5644 & 0.6720 \\
MS MARCO         & Politeness         & 0.4030 & 0.4840 & 0.5552 & 0.6638 \\
Natural Questions& RTT                & 0.6690 & 0.7556 & 0.8110 & 0.8878 \\
Natural Questions& Typos              & 0.6690 & 0.7556 & 0.8110 & 0.8878 \\
Natural Questions& Readability        & 0.6512 & 0.7300 & 0.7872 & 0.8614 \\
Natural Questions& Formality          & 0.6874 & 0.7700 & 0.8246 & 0.8944 \\
Natural Questions& Politeness         & 0.6538 & 0.7326 & 0.7860 & 0.8600 \\
PopQA            & RTT                & 0.6280 & 0.6952 & 0.7508 & 0.8344 \\
PopQA            & Typos              & 0.6280 & 0.6952 & 0.7508 & 0.8344 \\
PopQA            & Readability        & 0.6518 & 0.7168 & 0.7682 & 0.8432 \\
PopQA            & Formality          & 0.7408 & 0.7922 & 0.8316 & 0.8832 \\
PopQA            & Politeness         & 0.6744 & 0.7418 & 0.7962 & 0.8688 \\
\bottomrule
\end{tabular}
\caption{ModernBERT Retrieval performance (R@k) for original queries across datasets and linguistic modifications}
\end{table*}

\begin{table*}[ht]
\centering
\small
\setlength{\tabcolsep}{4pt}
\renewcommand{\arraystretch}{0.85}
\begin{tabular}{l l r r r r}
\toprule
\textbf{Dataset} & \textbf{Linguistics} & \textbf{R@5} & \textbf{R@10} & \textbf{R@20} & \textbf{R@100} \\
\midrule
PopQA            & Readability          & 0.5363 & 0.6148 & 0.6751 & 0.7796 \\
PopQA            & RTT                  & 0.4416 & 0.5090 & 0.5668 & 0.6856 \\
PopQA            & Typos                & 0.4868 & 0.5646 & 0.6242 & 0.7406 \\
PopQA            & Formality            & 0.6395 & 0.7109 & 0.7649 & 0.8493 \\
PopQA            & Politeness           & 0.6023 & 0.6776 & 0.7334 & 0.8317 \\
EntityQuestions & Readability          & 0.5325 & 0.5992 & 0.6607 & 0.7729 \\
EntityQuestions & RTT                  & 0.5632 & 0.6260 & 0.6784 & 0.7800 \\
EntityQuestions & Typos                & 0.5886 & 0.6518 & 0.7038 & 0.7918 \\
EntityQuestions & Formality            & 0.6251 & 0.6739 & 0.7156 & 0.7949 \\
EntityQuestions & Politeness           & 0.6502 & 0.7002 & 0.7436 & 0.8199 \\
MS MARCO         & Readability          & 0.3401 & 0.4291 & 0.5119 & 0.6461 \\
MS MARCO         & RTT                  & 0.2642 & 0.3340 & 0.3908 & 0.5146 \\
MS MARCO         & Typos                & 0.2810 & 0.3540 & 0.4188 & 0.5396 \\
MS MARCO         & Formality            & 0.3441 & 0.4311 & 0.5113 & 0.6479 \\
MS MARCO         & Politeness           & 0.3821 & 0.4637 & 0.5385 & 0.6579 \\
Natural Questions& Readability          & 0.5855 & 0.6741 & 0.7435 & 0.8393 \\
Natural Questions& RTT                  & 0.5470 & 0.6416 & 0.7192 & 0.8328 \\
Natural Questions& Typos                & 0.5790 & 0.6838 & 0.7512 & 0.8502 \\
Natural Questions& Formality            & 0.6220 & 0.7177 & 0.7890 & 0.8762 \\
Natural Questions& Politeness           & 0.6213 & 0.7054 & 0.7701 & 0.8539 \\
\bottomrule
\end{tabular}
\caption{ModernBERT Retrieval performance (R@k) for rewritten queries across datasets and linguistic modifications}
\end{table*}

\clearpage

\subsection{Scaling Number of documents}
\label{sec:scaling}

\begin{figure}[htbp]
    \centering
    \includegraphics[width=0.75\linewidth]{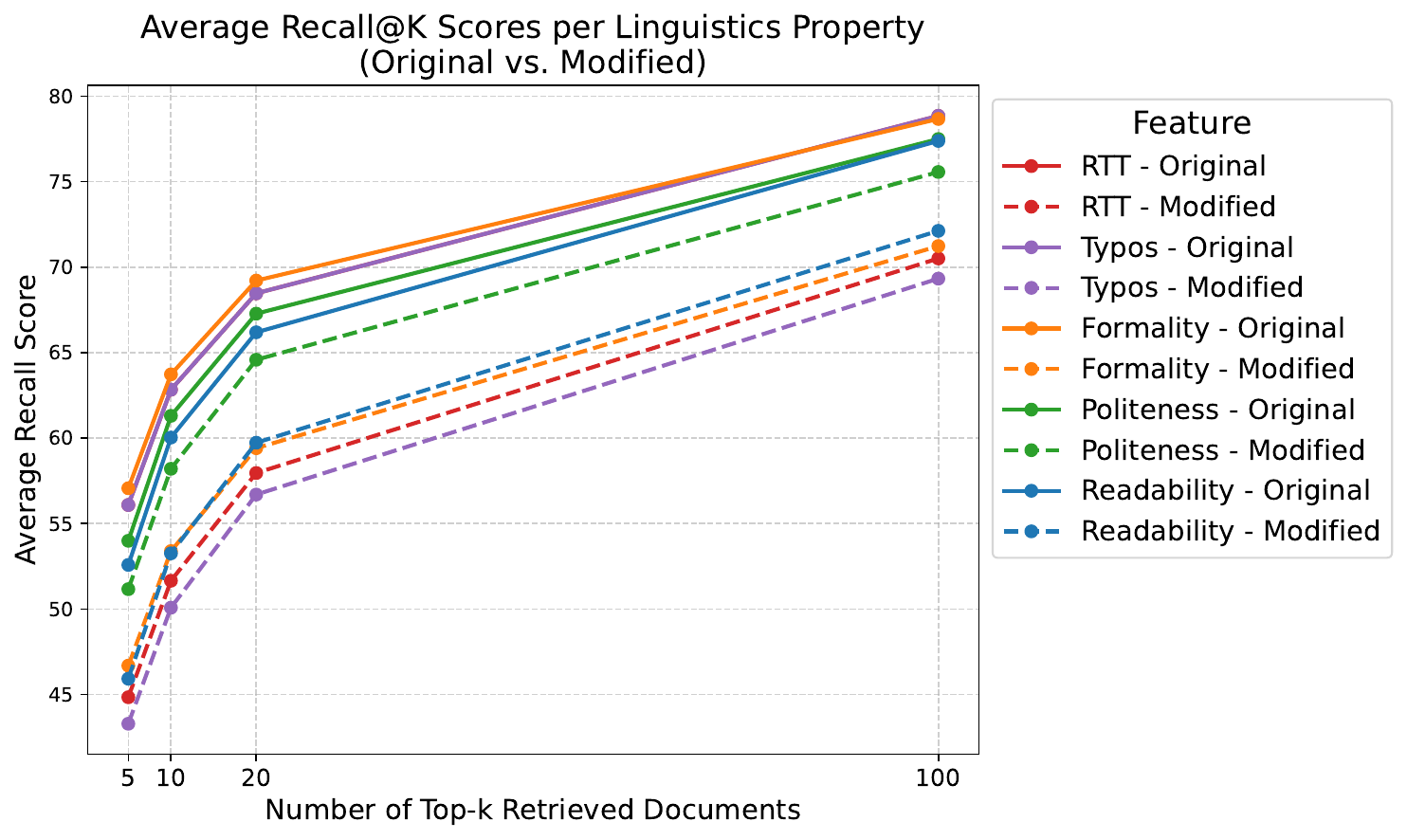}
    \caption{Average Recall@K increase as Number of Top-K Documents increases – Contriever.}
    \label{fig:avg_scores_contriever}
\end{figure}

\vspace{0.3cm}

\begin{figure}[htbp]
    \centering
    \includegraphics[width=0.75\linewidth]{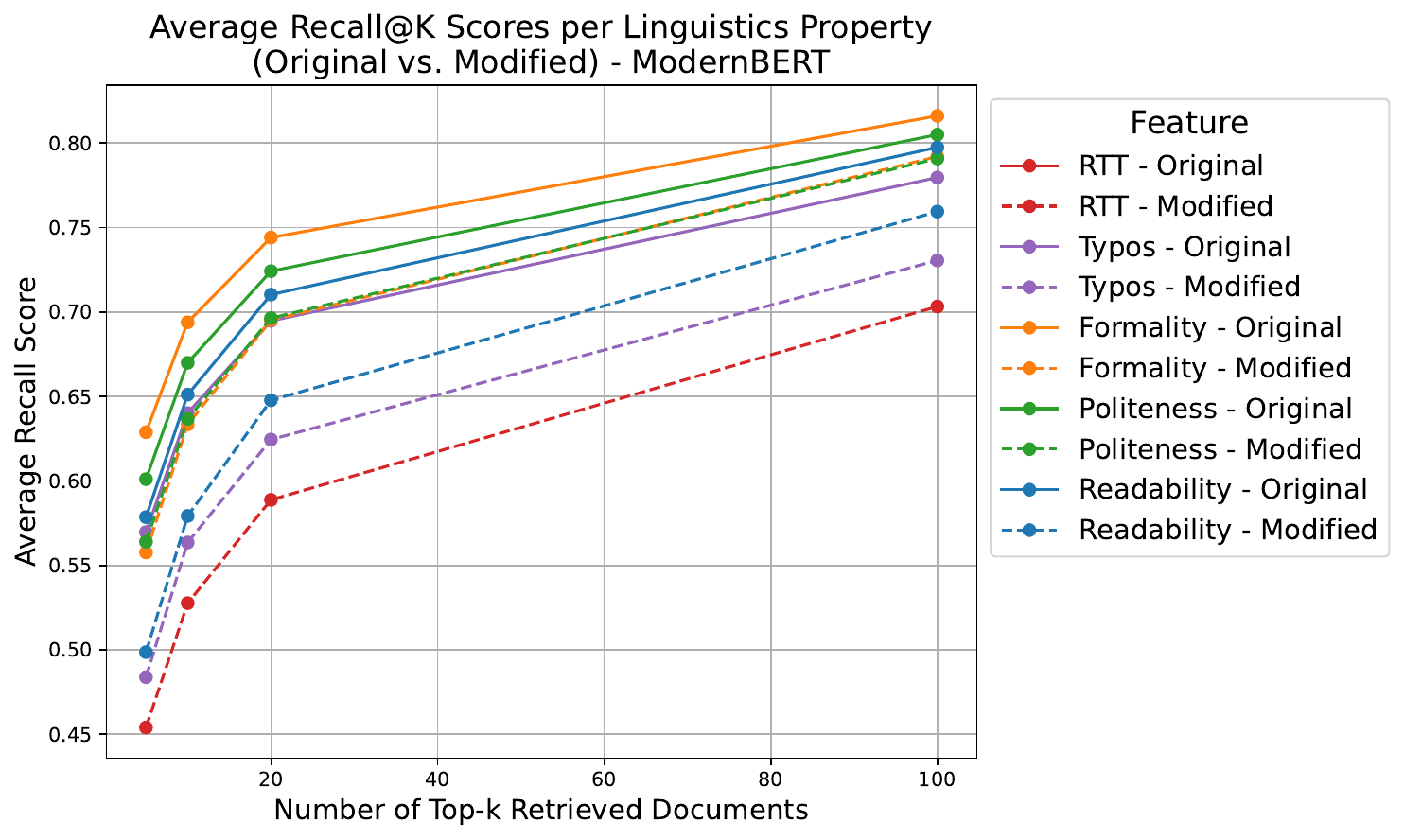}
    \caption{Average Recall@K increase as Number of Top-K Documents increases – ModernBERT.}
    \label{fig:avg_scores_modernbert}
\end{figure}

As mentioned in Section \ref{sec:retrieval}, the scaling of the number of documents decreases the degradation in performance, but does not mitigate the overall issue. As you can see in Figures \ref{fig:avg_scores_contriever} and \ref{fig:avg_scores_modernbert}, as K increases, higher recall benefits both original and rewritten queries, with performance gaps narrowing as correctly ranked documents appear at lower positions—suggesting linguistic variations primarily affect ranking order rather than complete retrieval failure. This hypothesis is confirmed in Section \ref{sec:reranker}, where the reranking shows considerable performance gains at R@5, showing that retrieval systems tend to push the correct documents for rewritten queries to lower ranks, and reranking helps prioritize them again, which is demonstrated by the reduction in performance degradation in Figure \ref{fig:hyde-rerank}.

\section{Full RAG Experiment Results}\label{app:full_rag_results}

\setlength{\tabcolsep}{3pt}
\begin{table*}[!htp]\centering
\resizebox{\textwidth}{!}{
\begin{tabular}{l rrr rrr rrr rrr}\toprule
\multicolumn{13}{c}{\textbf{Contriever, Formality, AM Score}} \\
\midrule
&\multicolumn{3}{c}{\textbf{PopQA}} & \multicolumn{3}{c}{\textbf{Natural Questions}} & \multicolumn{3}{c}{\textbf{MS MARCO}} & \multicolumn{3}{c}{\textbf{EntityQuestions}} \\
\cmidrule(lr){2-4}
\cmidrule(lr){5-7}
\cmidrule(lr){8-10}
\cmidrule(lr){11-13}
&Original &Rewritten &Delta &Original &Rewritten &Delta &Original &Rewritten &Delta &Original &Rewritten &Delta \\\midrule
gemma-2-2b-it &0.5858 &0.4719 &19.45\% &0.4432 &0.3742 &15.57\% &0.1416 &0.0955 &32.58\% &0.6026 &0.5446 &9.62\% \\
gemma-2-9b-it &0.6006 &0.4995 &16.83\% &0.4784 &0.4177 &12.68\% &0.1420 &0.0971 &31.64\% &0.6148 &0.5509 &10.39\% \\
gemma-2-27b-it &0.6126 &0.5094 &16.85\% &0.5046 &0.4447 &11.88\% &0.1384 &0.1025 &25.96\% &0.6300 &0.5691 &9.67\% \\
Llama-3.1-8B-Instruct &0.6122 &0.5098 &16.73\% &0.4930 &0.4376 &11.24\% &0.1580 &0.1121 &29.03\% &0.6156 &0.5621 &8.70\% \\
Llama-3.1-70B-Instruct &0.6386 &0.5297 &17.05\% &0.5170 &0.4331 &16.22\% &0.1586 &0.1057 &33.38\% &0.6352 &0.5747 &9.52\% \\
\bottomrule
\end{tabular}}
\caption{RAG experiment results with Contriever as retrieval model on AM scores.}
\end{table*}

\begin{table*}[!htp]\centering
\resizebox{\textwidth}{!}{
\begin{tabular}{l rrr rrr rrr rrr}
\toprule
\multicolumn{13}{c}{\textbf{Contriever, Formality, EM Score}} \\
\midrule
&\multicolumn{3}{c}{\textbf{PopQA}} & \multicolumn{3}{c}{\textbf{Natural Questions}} & \multicolumn{3}{c}{\textbf{MS MARCO}} & \multicolumn{3}{c}{\textbf{EntityQuestions}} \\
\cmidrule(lr){2-4}
\cmidrule(lr){5-7}
\cmidrule(lr){8-10}
\cmidrule(lr){11-13}
&Original &Rewritten &Delta &Original &Rewritten &Delta &Original &Rewritten &Delta &Original &Rewritten &Delta \\\midrule

gemma-2-2b-it &0.1454 &0.0151 &89.64\% &0.0550 &0.0205 &62.67\% &0.0196 &0.0058 &70.41\% &0.1914 &0.0330 &82.76\% \\
gemma-2-9b-it &0.0214 &0.0066 &69.16\% &0.0196 &0.0093 &52.72\% &0.0340 &0.0089 &73.92\% &0.0888 &0.0341 &61.64\% \\
gemma-2-27b-it &0.0044 &0.0122 &177.27\% &0.0188 &0.0147 &21.63\% &0.0058 &0.0037 &36.78\% &0.0292 &0.0505 &72.83\% \\
Llama-3.1-8B-Instruct &0.3282 &0.1348 &58.93\% &0.1482 &0.0525 &64.60\% &0.0216 &0.0055 &74.69\% &0.2896 &0.1429 &50.67\% \\
Llama-3.1-70B-Instruct &0.3432 &0.0884 &74.24\% &0.1234 &0.0369 &70.12\% &0.0244 &0.0069 &71.86\% &0.2996 &0.1177 &60.70\% \\
\bottomrule
\end{tabular}}
\caption{RAG experiment results with Contriever as retrieval model on EM scores.}
\end{table*}

\begin{table*}[!htp]\centering
\resizebox{\textwidth}{!}{
\begin{tabular}{lrrrrrrrrrrrrr}\toprule

\multicolumn{13}{c}{\textbf{Contriever, Formality, F1 Score}} \\
\midrule
&\multicolumn{3}{c}{\textbf{PopQA}} & \multicolumn{3}{c}{\textbf{Natural Questions}} & \multicolumn{3}{c}{\textbf{MS MARCO}} & \multicolumn{3}{c}{\textbf{EntityQuestions}} \\
\cmidrule(lr){2-4}
\cmidrule(lr){5-7}
\cmidrule(lr){8-10}
\cmidrule(lr){11-13}
&Original &Rewritten &Delta &Original &Rewritten &Delta &Original &Rewritten &Delta &Original &Rewritten &Delta \\\midrule

gemma-2-2b-it &0.3457 &0.2167 &37.31\% &0.2285 &0.1737 &23.99\% &0.2474 &0.1933 &21.86\% &0.3694 &0.2298 &37.79\% \\
gemma-2-9b-it &0.2809 &0.2274 &19.06\% &0.2142 &0.1911 &10.75\% &0.2614 &0.1988 &23.97\% &0.2996 &0.2436 &18.70\% \\
gemma-2-27b-it &0.2756 &0.2364 &14.24\% &0.2193 &0.1972 &10.10\% &0.2432 &0.2016 &17.10\% &0.2573 &0.2569 &0.12\% \\
Llama-3.1-8B-Instruct &0.4601 &0.2842 &38.22\% &0.3019 &0.1921 &36.37\% &0.2478 &0.1910 &22.93\% &0.4349 &0.2964 &31.85\% \\
Llama-3.1-70B-Instruct &0.4676 &0.2610 &44.17\% &0.2861 &0.1858 &35.06\% &0.2502 &0.1884 &24.71\% &0.4406 &0.2836 &35.65\% \\
\bottomrule
\end{tabular}}
\caption{RAG experiment results with Contriever as retrieval model on F1 scores.}
\end{table*}

\begin{table*}[!htp]\centering
\resizebox{\textwidth}{!}{
\begin{tabular}{lrrrrrrrrrrrrr}\toprule
\multicolumn{13}{c}{\textbf{ModernBERT, Formality, AM Score}} \\
\midrule
&\multicolumn{3}{c}{\textbf{PopQA}} & \multicolumn{3}{c}{\textbf{Natural Questions}} & \multicolumn{3}{c}{\textbf{MS MARCO}} & \multicolumn{3}{c}{\textbf{EntityQuestions}} \\
\cmidrule(lr){2-4}
\cmidrule(lr){5-7}
\cmidrule(lr){8-10}
\cmidrule(lr){11-13}
&Original &Rewritten &Delta &Original &Rewritten &Delta &Original &Rewritten &Delta &Original &Rewritten &Delta \\\midrule

gemma-2-2b-it &0.6126 &0.5291 &13.64\% &0.4608 &0.4035 &12.44\% &0.1690 &0.1296 &23.31\% &0.6012 &0.5551 &7.66\% \\
gemma-2-9b-it &0.6360 &0.5618 &11.67\% &0.4984 &0.4488 &9.95\% &0.1724 &0.1385 &19.68\% &0.6178 &0.5631 &8.85\% \\
gemma-2-27b-it &0.6508 &0.5713 &12.21\% &0.5230 &0.4711 &9.92\% &0.1676 &0.1355 &19.13\% &0.6264 &0.5813 &7.19\% \\
Llama-3.1-8B-Instruct &0.6488 &0.5695 &12.23\% &0.5228 &0.4679 &10.51\% &0.1982 &0.1633 &17.59\% &0.6158 &0.5717 &7.16\% \\
Llama-3.1-70B-Instruct &0.6722 &0.5950 &11.48\% &0.5378 &0.4728 &12.09\% &0.1986 &0.1609 &19.00\% &0.6350 &0.5893 &7.19\% \\
\bottomrule
\end{tabular}}
\caption{RAG experiment results with ModernBERT as retrieval model on AM scores.}
\end{table*}

\begin{table*}[!htp]\centering
\resizebox{\textwidth}{!}{
\begin{tabular}{lrrrrrrrrrrrrr}\toprule

\multicolumn{13}{c}{\textbf{ModernBERT, Formality, EM Score}} \\
\midrule
&\multicolumn{3}{c}{\textbf{PopQA}} & \multicolumn{3}{c}{\textbf{Natural Questions}} & \multicolumn{3}{c}{\textbf{MS MARCO}} & \multicolumn{3}{c}{\textbf{EntityQuestions}} \\
\cmidrule(lr){2-4}
\cmidrule(lr){5-7}
\cmidrule(lr){8-10}
\cmidrule(lr){11-13}
&Original &Rewritten &Delta &Original &Rewritten &Delta &Original &Rewritten &Delta &Original &Rewritten &Delta \\\midrule

gemma-2-2b-it &0.1554 &0.0167 &89.23\% &0.0558 &0.0204 &63.44\% &0.0220 &0.0066 &70.00\% &0.1964 &0.0299 &84.76\% \\
gemma-2-9b-it &0.0282 &0.0072 &74.47\% &0.0210 &0.0091 &56.83\% &0.0358 &0.0096 &73.18\% &0.0992 &0.0356 &64.11\% \\
gemma-2-27b-it &0.0052 &0.0105 &101.28\% &0.0202 &0.0141 &30.36\% &0.0096 &0.0039 &59.03\% &0.0320 &0.0496 &55.00\% \\
Llama-3.1-8B-Instruct &0.3500 &0.1525 &56.42\% &0.1642 &0.0551 &66.42\% &0.0286 &0.0115 &59.67\% &0.2886 &0.1429 &50.47\% \\
Llama-3.1-70B-Instruct &0.3748 &0.1008 &73.11\% &0.1342 &0.0375 &72.03\% &0.0324 &0.0123 &61.93\% &0.3088 &0.1203 &61.03\% \\
\bottomrule
\end{tabular}}
\caption{RAG experiment results with ModernBERT as retrieval model on EM scores.}
\end{table*}

\begin{table*}[!htp]\centering
\resizebox{\textwidth}{!}{
\begin{tabular}{lrrrrrrrrrrrrr}\toprule

\multicolumn{13}{c}{\textbf{ModernBERT, Formality, F1 Score}} \\
\midrule
&\multicolumn{3}{c}{\textbf{PopQA}} & \multicolumn{3}{c}{\textbf{Natural Questions}} & \multicolumn{3}{c}{\textbf{MS MARCO}} & \multicolumn{3}{c}{\textbf{EntityQuestions}} \\
\cmidrule(lr){2-4}
\cmidrule(lr){5-7}
\cmidrule(lr){8-10}
\cmidrule(lr){11-13}
&Original &Rewritten &Delta &Original &Rewritten &Delta &Original &Rewritten &Delta &Original &Rewritten &Delta \\\midrule

gemma-2-2b-it &0.3666 &0.2525 &31.14\% &0.2346 &0.1833 &21.88\% &0.2800 &0.2276 &18.73\% &0.3696 &0.2327 &37.05\% \\
gemma-2-9b-it &0.3018 &0.2642 &12.48\% &0.2217 &0.2007 &9.46\% &0.2927 &0.2429 &17.01\% &0.3082 &0.2511 &18.55\% \\
gemma-2-27b-it &0.2965 &0.2718 &8.33\% &0.2273 &0.2053 &9.67\% &0.2746 &0.2416 &12.03\% &0.2574 &0.2625 &1.96\% \\
Llama-3.1-8B-Instruct &0.4871 &0.3223 &33.84\% &0.3205 &0.2005 &37.43\% &0.2880 &0.2396 &16.81\% &0.4291 &0.2985 &30.45\% \\
Llama-3.1-70B-Instruct &0.4998 &0.2968 &40.63\% &0.2993 &0.1972 &34.10\% &0.2931 &0.2467 &15.83\% &0.4452 &0.2891 &35.07\% \\
\bottomrule
\end{tabular}}
\caption{RAG experiment results with ModernBERT as retrieval model on F1 scores.}
\end{table*}

\clearpage

\section{Full Query Expansion Experiment Results}\label{app:hyde}

The following tables provide results of Retrieval and Generation components across MS MARCO and PopQA.

\begin{table}[ht]
\centering
\scriptsize
\setlength{\tabcolsep}{4pt}
\renewcommand{\arraystretch}{0.85}
\begin{tabular}{l l r r r r}
\toprule
\textbf{Dataset} & \textbf{Linguistics} & \textbf{R@5} & \textbf{R@10} & \textbf{R@20} & \textbf{R@100} \\
\midrule
MS MARCO & RTT           & 37.10 & 43.74 & 50.56 & 63.76 \\
MS MARCO & Typos         & 37.02 & 43.66 & 50.40 & 63.72 \\
MS MARCO & Readability   & 29.22 & 37.34 & 45.60 & 61.00 \\
MS MARCO & Formality     & 28.92 & 36.88 & 45.16 & 60.50 \\
MS MARCO & Politeness    & 29.44 & 37.54 & 45.68 & 60.52 \\
PopQA & RTT              & 64.50 & 69.72 & 74.18 & 83.32 \\
PopQA & Typos            & 64.42 & 69.60 & 73.82 & 83.36 \\
PopQA & Readability      & 64.30 & 69.62 & 74.70 & 84.28 \\
PopQA & Formality        & 72.82 & 77.68 & 81.30 & 88.00 \\
PopQA & Politeness       & 66.74 & 71.62 & 76.50 & 85.94 \\
\bottomrule
\end{tabular}
\caption{Contriever retrieval results with RAG+HyDE for original queries}
\end{table}

\begin{table}[ht]
\centering
\scriptsize
\setlength{\tabcolsep}{4pt}
\renewcommand{\arraystretch}{0.85}
\begin{tabular}{l l r r r r}
\toprule
\textbf{Dataset} & \textbf{Linguistics} & \textbf{R@5} & \textbf{R@10} & \textbf{R@20} & \textbf{R@100} \\
\midrule
MS MARCO & RTT           & 32.80 & 38.74 & 44.46 & 57.70 \\
MS MARCO & Typos         & 36.00 & 42.10 & 49.36 & 62.84 \\
MS MARCO & Readability   & 26.14 & 34.88 & 42.74 & 58.52 \\
MS MARCO & Formality     & 25.66 & 33.06 & 41.38 & 58.60 \\
MS MARCO & Politeness    & 28.44 & 36.28 & 44.58 & 59.44 \\
PopQA & RTT              & 44.62 & 49.06 & 53.80 & 65.72 \\
PopQA & Typos            & 59.72 & 64.58 & 69.50 & 80.30 \\
PopQA & Readability      & 59.00 & 63.96 & 68.72 & 79.34 \\
PopQA & Formality        & 68.60 & 72.82 & 76.42 & 84.20 \\
PopQA & Politeness       & 65.18 & 70.06 & 74.60 & 83.70 \\
\bottomrule
\end{tabular}
\caption{Contriever retrieval results with RAG+HyDE for rewritten queries}
\end{table}

\begin{table}[ht]
\centering
\scriptsize
\setlength{\tabcolsep}{4pt}
\renewcommand{\arraystretch}{0.85}
\begin{tabular}{l l r r r r}
\toprule
\textbf{Dataset} & \textbf{Linguistics} & \textbf{R@5} & \textbf{R@10} & \textbf{R@20} & \textbf{R@100} \\
\midrule
PopQA & RTT           & 0.5984 & 0.6496 & 0.6980 & 0.7936 \\
PopQA & Typos         & 0.5928 & 0.6444 & 0.6962 & 0.7970 \\
PopQA & Readability   & 0.6176 & 0.6720 & 0.7170 & 0.8164 \\
PopQA & Formality     & 0.7064 & 0.7518 & 0.7938 & 0.8670 \\
PopQA & Politeness    & 0.6322 & 0.6780 & 0.7272 & 0.8278 \\
MS MARCO & RTT           & 0.2994 & 0.3672 & 0.4362 & 0.5546 \\
MS MARCO & Typos         & 0.3030 & 0.3698 & 0.4376 & 0.5560 \\
MS MARCO & Readability   & 0.3830 & 0.4616 & 0.5414 & 0.6608 \\
MS MARCO & Formality     & 0.3816 & 0.4594 & 0.5380 & 0.6520 \\
MS MARCO & Politeness    & 0.3800 & 0.4580 & 0.5314 & 0.6490 \\
\bottomrule
\end{tabular}
\caption{ModernBERT retrieval results with RAG+HyDE for original queries}
\end{table}

\begin{table}[hb]
\centering
\scriptsize
\setlength{\tabcolsep}{4pt}
\renewcommand{\arraystretch}{0.85}
\begin{tabular}{l l r r r r}
\toprule
\textbf{Dataset} & \textbf{Linguistics} & \textbf{R@5} & \textbf{R@10} & \textbf{R@20} & \textbf{R@100} \\
\midrule
PopQA & RTT           & 0.3750 & 0.4230 & 0.4678 & 0.5802 \\
PopQA & Typos         & 0.5388 & 0.5940 & 0.6538 & 0.7584 \\
PopQA & Readability   & 0.5468 & 0.5904 & 0.6394 & 0.7402 \\
PopQA & Formality     & 0.6520 & 0.6970 & 0.7404 & 0.8228 \\
PopQA & Politeness    & 0.6148 & 0.6640 & 0.7094 & 0.8064 \\
MS MARCO & RTT           & 0.2438 & 0.3020 & 0.3694 & 0.4986 \\
MS MARCO & Typos         & 0.2876 & 0.3546 & 0.4280 & 0.5438 \\
MS MARCO & Readability   & 0.3462 & 0.4326 & 0.5172 & 0.6480 \\
MS MARCO & Formality     & 0.3532 & 0.4376 & 0.5122 & 0.6402 \\
MS MARCO & Politeness    & 0.3724 & 0.4510 & 0.5232 & 0.6454 \\
\bottomrule
\end{tabular}
\caption{ModernBERT retrieval results with RAG+HyDE for rewritten queries}
\end{table}

\begin{table*}[!htp]\centering
\small
\begin{tabular}{lrrrrrr}\toprule
Dataset &Linguistics &LLMs &Original &Rewritten &Delta \\\midrule
MS MARCO &RTT &Llama-3.1-8B-Instruct &0.1466 &0.1014 &30.83\% \\
MS MARCO &RTT &Qwen2.5-7B-Instruct &0.1418 &0.1050 &25.95\% \\
MS MARCO &RTT &gemma-2-9b-it &0.1366 &0.0966 &29.28\% \\
MS MARCO &Typos &Llama-3.1-8B-Instruct &0.1466 &0.1378 &6.00\% \\
MS MARCO &Typos &Qwen2.5-7B-Instruct &0.1418 &0.1328 &6.35\% \\
MS MARCO &Typos &gemma-2-9b-it &0.1366 &0.1284 &6.00\% \\
MS MARCO &Formality &Llama-3.1-8B-Instruct &0.1626 &0.1430 &12.05\% \\
\bottomrule
\end{tabular}
\caption{RAG experiment results with query expansion and Contriever as retrieval model on AM scores.}
\end{table*}

\begin{table*}[!htp]\centering
\small
\begin{tabular}{lrrrrrr}\toprule
Dataset &Linguistics &LLMs &Original &Rewritten &Delta \\\midrule
MS MARCO &RTT &Llama-3.1-8B-Instruct &0.0318 &0.0166 &47.80\% \\
MS MARCO &RTT &Qwen2.5-7B-Instruct &0.0358 &0.0244 &31.84\% \\
MS MARCO &RTT &gemma-2-9b-it &0.0350 &0.0254 &27.43\% \\
MS MARCO &Typos &Llama-3.1-8B-Instruct &0.0318 &0.0286 &10.06\% \\
MS MARCO &Typos &Qwen2.5-7B-Instruct &0.0358 &0.0294 &17.88\% \\
MS MARCO &Typos &gemma-2-9b-it &0.0350 &0.0314 &10.29\% \\
MS MARCO &Formality &Llama-3.1-8B-Instruct &0.0234 &0.0098 &58.12\% \\
\bottomrule
\end{tabular}
\caption{RAG experiment results with query expansion and Contriever as retrieval model on EM scores.}
\end{table*}

\begin{table*}[!htp]\centering
\small
\begin{tabular}{lrrrrrr}\toprule
Dataset &Linguistics &LLMs &Original &Rewritten &Delta \\\midrule
MS MARCO &RTT &Llama-3.1-8B-Instruct &0.2412 &0.1894 &21.49\% \\
MS MARCO &RTT &Qwen2.5-7B-Instruct &0.2453 &0.2037 &16.93\% \\
MS MARCO &RTT &gemma-2-9b-it &0.2458 &0.1995 &18.81\% \\
MS MARCO &Typos &Llama-3.1-8B-Instruct &0.2412 &0.2323 &3.71\% \\
MS MARCO &Typos &Qwen2.5-7B-Instruct &0.2453 &0.2357 &3.89\% \\
MS MARCO &Typos &gemma-2-9b-it &0.2458 &0.2381 &3.12\% \\
MS MARCO &Formality &Llama-3.1-8B-Instruct &0.2593 &0.2270 &12.44\% \\
\bottomrule
\end{tabular}
\caption{RAG experiment results with query expansion and Contriever as retrieval model on F1 scores.}
\end{table*}

\begin{table*}[!htp]\centering
\small
\begin{tabular}{lrrrrrr}\toprule
Dataset &Linguistics &LLMs &Original &Rewritten &Delta \\\midrule
MS MARCO &RTT &Llama-3.1-8B-Instruct &0.1706 &0.1222 &28.37\% \\
MS MARCO &RTT &Qwen2.5-7B-Instruct &0.1634 &0.1216 &25.58\% \\
MS MARCO &RTT &gemma-2-9b-it &0.1560 &0.1138 &27.05\% \\
MS MARCO &Typos &Llama-3.1-8B-Instruct &0.1746 &0.1642 &5.96\% \\
MS MARCO &Typos &Qwen2.5-7B-Instruct &0.1592 &0.1522 &4.40\% \\
MS MARCO &Typos &gemma-2-9b-it &0.1578 &0.1462 &7.35\% \\
MS MARCO &Formality &Llama-3.1-8B-Instruct &0.1896 &0.1710 &9.81\% \\
\bottomrule
\end{tabular}
\caption{RAG experiment results with query expansion and ModernBERT as retrieval model on AM scores.}
\end{table*}

\begin{table*}[!htp]\centering
\small
\begin{tabular}{lrrrrrr}\toprule
Dataset &Linguistics &LLMs &Original &Rewritten &Delta \\\midrule
MS MARCO &RTT &Llama-3.1-8B-Instruct &0.0292 &0.0218 &25.34\% \\
MS MARCO &RTT &Qwen2.5-7B-Instruct &0.0380 &0.0348 &8.42\% \\
MS MARCO &RTT &gemma-2-9b-it &0.0392 &0.0270 &31.12\% \\
MS MARCO &Typos &Llama-3.1-8B-Instruct &0.0358 &0.0316 &11.73\% \\
MS MARCO &Typos &Qwen2.5-7B-Instruct &0.0416 &0.0322 &22.60\% \\
MS MARCO &Typos &gemma-2-9b-it &0.0400 &0.0334 &16.50\% \\
MS MARCO &Formality &Llama-3.1-8B-Instruct &0.0288 &0.0142 &50.69\% \\
\bottomrule
\end{tabular}
\caption{RAG experiment results with query expansion and ModernBERT as retrieval model on EM scores.}
\end{table*}

\begin{table*}[!htp]\centering
\small
\begin{tabular}{lrrrrrr}\toprule
Dataset &Linguistics &LLMs &Original &Rewritten &Delta \\\midrule
MS MARCO &RTT &Llama-3.1-8B-Instruct &0.2593 &0.2080 &19.76\% \\
MS MARCO &RTT &Qwen2.5-7B-Instruct &0.2697 &0.2247 &16.69\% \\
MS MARCO &RTT &gemma-2-9b-it &0.2669 &0.2176 &18.48\% \\
MS MARCO &Typos &Llama-3.1-8B-Instruct &0.2667 &0.2544 &4.62\% \\
MS MARCO &Typos &Qwen2.5-7B-Instruct &0.2717 &0.2583 &4.95\% \\
MS MARCO &Typos &gemma-2-9b-it &0.2677 &0.2540 &5.12\% \\
MS MARCO &Formality &Llama-3.1-8B-Instruct &0.2855 &0.2559 &10.36\% \\
\bottomrule
\end{tabular}
\caption{RAG experiment results with query expansion and ModernBERT as retrieval model on F1 scores.}
\end{table*}

\clearpage

\section{Full Reranking Experiment Results}\label{app:rerank}
%retrieval_section

\begin{table}[ht]
\centering
\scriptsize
\setlength{\tabcolsep}{4pt}
\renewcommand{\arraystretch}{0.85}
\begin{tabular}{l l r r r r}
\toprule
\textbf{Dataset} & \textbf{Linguistics} & \textbf{R@5} & \textbf{R@10} & \textbf{R@20} & \textbf{R@100} \\
\midrule
MS MARCO & RTT          & 0.4620 & 0.5208 & 0.5618 & 0.6102 \\
MS MARCO & Typos        & 0.4620 & 0.5208 & 0.5618 & 0.6102 \\
MS MARCO & Readability  & 0.4058 & 0.4688 & 0.5188 & 0.5718 \\
MS MARCO & Formality    & 0.4070 & 0.4714 & 0.5154 & 0.5598 \\
MS MARCO & Politeness   & 0.4078 & 0.4706 & 0.5146 & 0.5664 \\
PopQA    & RTT          & 0.6640 & 0.7230 & 0.7674 & 0.8192 \\
PopQA    & Typos        & 0.6640 & 0.7230 & 0.7674 & 0.8192 \\
PopQA    & Readability  & 0.6864 & 0.7506 & 0.7870 & 0.8344 \\
PopQA    & Formality    & 0.7620 & 0.8100 & 0.8404 & 0.8760 \\
PopQA    & Politeness   & 0.7050 & 0.7722 & 0.8096 & 0.8534 \\
\bottomrule
\end{tabular}
\caption{Contriever retrieval results with RAG+Reranking for original queries}
\end{table}

\begin{table}[ht]
\centering
\scriptsize
\setlength{\tabcolsep}{4pt}
\renewcommand{\arraystretch}{0.85}
\begin{tabular}{l l r r r r}
\toprule
\textbf{Dataset} & \textbf{Linguistics} & \textbf{R@5} & \textbf{R@10} & \textbf{R@20} & \textbf{R@100} \\
\midrule
MS MARCO & RTT          & 0.4144 & 0.4674 & 0.5158 & 0.5754 \\
MS MARCO & Typos        & 0.4106 & 0.4598 & 0.4932 & 0.5440 \\
MS MARCO & Readability  & 0.3061 & 0.3773 & 0.4420 & 0.5235 \\
MS MARCO & Formality    & 0.2755 & 0.3341 & 0.3803 & 0.4365 \\
MS MARCO & Politeness   & 0.3665 & 0.4304 & 0.4751 & 0.5448 \\
PopQA    & RTT          & 0.4900 & 0.5510 & 0.5946 & 0.6654 \\
PopQA    & Typos        & 0.5420 & 0.5888 & 0.6238 & 0.6724 \\
PopQA    & Readability  & 0.5839 & 0.6421 & 0.6835 & 0.7332 \\
PopQA    & Formality    & 0.6837 & 0.7266 & 0.7543 & 0.7856 \\
PopQA    & Politeness   & 0.6655 & 0.7300 & 0.7701 & 0.8081 \\
\bottomrule
\end{tabular}
\caption{Contriever retrieval results with RAG+Reranking for rewritten queries}
\end{table}

\begin{table}[ht]
\centering
\scriptsize
\setlength{\tabcolsep}{4pt}
\renewcommand{\arraystretch}{0.85}
\begin{tabular}{l l r r r r}
\toprule
\textbf{Dataset} & \textbf{Linguistics} & \textbf{R@5} & \textbf{R@10} & \textbf{R@20} & \textbf{R@100} \\
\midrule
PopQA     & RTT            & 0.6836 & 0.7462 & 0.7934 & 0.8344 \\
PopQA     & Typos          & 0.6836 & 0.7462 & 0.7934 & 0.8344 \\
PopQA     & Readability    & 0.7012 & 0.7620 & 0.8024 & 0.8432 \\
PopQA     & Formality      & 0.7684 & 0.8162 & 0.8502 & 0.8832 \\
PopQA     & Politeness     & 0.7282 & 0.7878 & 0.8310 & 0.8688 \\
MS MARCO  & RTT            & 0.3628 & 0.4302 & 0.4860 & 0.5680 \\
MS MARCO  & Typos          & 0.3628 & 0.4302 & 0.4860 & 0.5680 \\
MS MARCO  & Readability    & 0.4428 & 0.5194 & 0.5876 & 0.6746 \\
MS MARCO  & Formality      & 0.4482 & 0.5284 & 0.5932 & 0.6720 \\
MS MARCO  & Politeness     & 0.4478 & 0.5222 & 0.5802 & 0.6638 \\
\bottomrule
\end{tabular}
\caption{ModernBERT retrieval results with RAG+Reranking for original queries}
\end{table}

\begin{table}[ht]
\centering
\scriptsize
\setlength{\tabcolsep}{4pt}
\renewcommand{\arraystretch}{0.85}
\begin{tabular}{l l r r r r}
\toprule
\textbf{Dataset} & \textbf{Linguistics} & \textbf{R@5} & \textbf{R@10} & \textbf{R@20} & \textbf{R@100} \\
\midrule
PopQA     & RTT            & 0.5096 & 0.5744 & 0.6204 & 0.6856 \\
PopQA     & Typos          & 0.5940 & 0.6496 & 0.6886 & 0.7406 \\
PopQA     & Readability    & 0.6237 & 0.6836 & 0.7289 & 0.7796 \\
PopQA     & Formality      & 0.7337 & 0.7809 & 0.8113 & 0.8493 \\
PopQA     & Politeness     & 0.6888 & 0.7487 & 0.7860 & 0.8317 \\
MS MARCO  & RTT            & 0.2940 & 0.3558 & 0.4160 & 0.5146 \\
MS MARCO  & Typos          & 0.3248 & 0.3902 & 0.4548 & 0.5396 \\
MS MARCO  & Readability    & 0.3279 & 0.4179 & 0.4994 & 0.6461 \\
MS MARCO  & Formality      & 0.3479 & 0.4392 & 0.5224 & 0.6479 \\
MS MARCO  & Politeness     & 0.4060 & 0.4882 & 0.5553 & 0.6579 \\
\bottomrule
\end{tabular}
\caption{ModernBERT retrieval results with RAG+Reranking for rewritten queries}
\end{table}

%generation_section

\begin{table*}[!htp]\centering
\small
\begin{tabular}{lrrrrrr}\toprule
Dataset &Linguistics &LLMs &Original &Rewritten &Delta \\\midrule
MS MARCO &RTT &Llama-3.1-8B-Instruct &0.1428 &0.1302 &8.82\% \\
MS MARCO &RTT &Qwen2.5-7B-Instruct &0.1416 &0.1290 &8.90\% \\
MS MARCO &RTT &gemma-2-9b-it &0.1292 &0.1202 &6.97\% \\
MS MARCO &Typos &Llama-3.1-8B-Instruct &0.1428 &0.1118 &21.71\% \\
MS MARCO &Typos &Qwen2.5-7B-Instruct &0.1416 &0.1120 &20.90\% \\
MS MARCO &Typos &gemma-2-9b-it &0.1292 &0.1066 &17.49\% \\
MS MARCO &Formality &Llama-3.1-8B-Instruct &0.1552 &0.1127 &27.36\% \\
\bottomrule
\end{tabular}
\caption{Generation results with Contriever retrieval and reranking on AM scores.}
\end{table*}

\begin{table*}[!htp]\centering
\small
\begin{tabular}{lrrrrrr}\toprule
Dataset &Linguistics &LLMs &Original &Rewritten &Delta \\\midrule
MS MARCO &RTT &Llama-3.1-8B-Instruct &0.0288 &0.0214 &25.69\% \\
MS MARCO &RTT &Qwen2.5-7B-Instruct &0.0336 &0.0302 &10.12\% \\
MS MARCO &RTT &gemma-2-9b-it &0.0388 &0.0288 &25.77\% \\
MS MARCO &Typos &Llama-3.1-8B-Instruct &0.0288 &0.0174 &39.58\% \\
MS MARCO &Typos &Qwen2.5-7B-Instruct &0.0336 &0.0254 &24.40\% \\
MS MARCO &Typos &gemma-2-9b-it &0.0388 &0.0308 &20.62\% \\
MS MARCO &Formality &Llama-3.1-8B-Instruct &0.0232 &0.0059 &74.43\% \\
\bottomrule
\end{tabular}
\caption{Generation results with Contriever retrieval and reranking on EM scores.}
\end{table*}

\begin{table*}[!htp]\centering
\small
\begin{tabular}{lrrrrrr}\toprule
Dataset &Linguistics &LLMs &Original &Rewritten &Delta \\\midrule
MS MARCO &RTT &Llama-3.1-8B-Instruct &0.2310 &0.2146 &7.12\% \\
MS MARCO &RTT &Qwen2.5-7B-Instruct &0.2402 &0.2272 &5.42\% \\
MS MARCO &RTT &gemma-2-9b-it &0.2394 &0.2211 &7.64\% \\
MS MARCO &Typos &Llama-3.1-8B-Instruct &0.2310 &0.2015 &12.76\% \\
MS MARCO &Typos &Qwen2.5-7B-Instruct &0.2402 &0.2090 &12.99\% \\
MS MARCO &Typos &gemma-2-9b-it &0.2394 &0.2124 &11.26\% \\
MS MARCO &Formality &Llama-3.1-8B-Instruct &0.2493 &0.1913 &23.26\% \\
\bottomrule
\end{tabular}
\caption{Generation results with Contriever retrieval and reranking on F1 scores.}
\end{table*}

\begin{table*}[!htp]\centering
\small
\begin{tabular}{lrrrrrr}\toprule
Dataset &Linguistics &LLMs &Original &Rewritten &Delta \\\midrule
MS MARCO &RTT &Llama-3.1-8B-Instruct &0.1964 &0.1390 &29.23\% \\
MS MARCO &RTT &Qwen2.5-7B-Instruct &0.1882 &0.1354 &28.06\% \\
MS MARCO &RTT &gemma-2-9b-it &0.1690 &0.1246 &26.27\% \\
MS MARCO &Typos &Llama-3.1-8B-Instruct &0.1964 &0.1772 &9.78\% \\
MS MARCO &Typos &Qwen2.5-7B-Instruct &0.1882 &0.1678 &10.84\% \\
MS MARCO &Typos &gemma-2-9b-it &0.1690 &0.1618 &4.26\% \\
MS MARCO &Formality &Llama-3.1-8B-Instruct &0.2118 &0.1655 &21.84\% \\
\bottomrule
\end{tabular}
\caption{Generation results with ModernBERT retrieval and reranking on AM scores.}
\end{table*}

\begin{table*}[!htp]\centering
\small
\begin{tabular}{lrrrrrr}\toprule
Dataset &Linguistics &LLMs &Original &Rewritten &Delta \\\midrule
MS MARCO &RTT &Llama-3.1-8B-Instruct &0.0446 &0.0248 &44.39\% \\
MS MARCO &RTT &Qwen2.5-7B-Instruct &0.0472 &0.0370 &21.61\% \\
MS MARCO &RTT &gemma-2-9b-it &0.0428 &0.0326 &23.83\% \\
MS MARCO &Typos &Llama-3.1-8B-Instruct &0.0446 &0.0342 &23.32\% \\
MS MARCO &Typos &Qwen2.5-7B-Instruct &0.0472 &0.0376 &20.34\% \\
MS MARCO &Typos &gemma-2-9b-it &0.0428 &0.0352 &17.76\% \\
MS MARCO &Formality &Llama-3.1-8B-Instruct &0.0332 &0.0105 &68.27\% \\
\bottomrule
\end{tabular}
\caption{Generation results with ModernBERT retrieval and reranking on EM scores.}
\end{table*}

\begin{table*}[!htp]\centering
\small
\begin{tabular}{lrrrrrr}\toprule
Dataset &Linguistics &LLMs &Original &Rewritten &Delta \\\midrule
MS MARCO &RTT &Llama-3.1-8B-Instruct &0.2837 &0.2203 &22.35\% \\
MS MARCO &RTT &Qwen2.5-7B-Instruct &0.2866 &0.2349 &18.04\% \\
MS MARCO &RTT &gemma-2-9b-it &0.2797 &0.2308 &17.49\% \\
MS MARCO &Typos &Llama-3.1-8B-Instruct &0.2837 &0.2622 &7.56\% \\
MS MARCO &Typos &Qwen2.5-7B-Instruct &0.2866 &0.2672 &6.78\% \\
MS MARCO &Typos &gemma-2-9b-it &0.2797 &0.2638 &5.68\% \\
MS MARCO &Formality &Llama-3.1-8B-Instruct &0.3004 &0.2395 &20.25\% \\
\bottomrule
\end{tabular}
\caption{Generation results with ModernBERT retrieval and reranking on F1 scores.}
\end{table*}

\clearpage
\section{Rewriting Prompts}
\label{app: rewriting_prompts}
\subsection{Formality}

\begin{description}

  \item[Prompt 1:] \hfill \\
  \begingroup\ttfamily
  You are an AI assistant skilled at transforming formal queries into casual, everyday language. 
  Rewrite the following query so that it sounds very informal. Experiment with different colloquial 
  openings, varied sentence constructions, and a mix of slang, idioms, and casual expressions 
  throughout the sentence. Avoid using the same phrase repeatedly (e.g., "hey, so like") and ensure 
  the meaning remains unchanged.
  \par\endgroup

  \item[Prompt 2:] \hfill \\
  \begingroup\ttfamily
  Your task is to convert the given query into an informal version that feels natural and conversational. 
  Instead of a uniform introductory phrase, use a range of informal expressions (such as interjections, 
  casual questions, or slang) at different parts of the sentence. Mix up the structure—sometimes start with 
  an interjection, other times rephrase the sentence completely—while keeping the original meaning intact.
  \par\endgroup

  \item[Prompt 3:] \hfill \\
  \begingroup\ttfamily
  Task: Transform Formal to Extremely Informal Language

  Convert the following formal sentence into an extremely informal, messy, and natural version. 
  The output should sound like authentic, real-world casual speech—as if spoken in an informal chat, 
  online conversation, or street talk.

  Critical Rules:
  - Diversity is key: No two sentences should follow the same pattern.
  - Break formal sentence structures: Chop up long phrases, reorder words, or make them flow casually.
  - Use a wide range of informality techniques—DO NOT rely on just contractions or slang.
  - Avoid starting every sentence with the same phrase or discourse marker.
  - Embrace randomness: Let the outputs sound wild, real, and unpredictable.
  - Do not start your sentences with "Yo", "Hey so like", etc.

  Final Execution Instruction:  
  Generate an informal version of the following sentence that:  
  - Uses multiple different informality techniques.  
  - Avoids repetitive sentence structures or patterns.  
  - Sounds raw, conversational, and unpredictable.

  Original Query:
  \par\endgroup

\end{description}

\subsection{Readablility}

\begin{description}

  \item[Prompt 1:] \hfill \\
  \begingroup\ttfamily
1. Task Definition:

You are rewriting a query to make it significantly less readable while preserving the original semantic meaning as closely as possible.

2. Constraints \& Goals:

- Flesch Reading Ease Score: The rewritten text must have a Flesch score below 60 (preferably below 50).

- Semantic Similarity: The rewritten text must have SBERT similarity > 0.7 compared with the original query.

- Length: The rewritten text must remain approximately the same length as the original query (±10\%).

- Preserve Domain Terminology: Do not remove or drastically change domain-specific words, abbreviations, or technical terms (e.g., "IRS," "distance," etc.). 

- Abbreviation: Do not expand abbreviations unless the original query already used the expanded form.

- No New Information: You must not add additional details beyond what the original query states.

- Question Format: Retain the form of a question if the original is posed as a question.

3. How to Increase Complexity:

- Lexical Changes: Use advanced or academic synonyms only for common words. For domain or key terms (e.g., "distance," "IRS," "tax"), keep the original term or use a very close synonym if necessary to maintain meaning.

- Syntactic Complexity: Introduce passive voice, nominalizations, embedded clauses, and parenthetical or subordinate phrases. Ensure the sentence flow is more formal and convoluted without changing the core meaning.

- Redundancy \& Formality: Employ circumlocution and excessively formal expressions (e.g., "due to the fact that" instead of "because") while avoiding any semantic drift.

- Dense, Indirect Construction: Favor longer phrases, indirect references, and wordiness. Avoid direct or simple phrasing.  

Original Query: ...

Less Readable Query:

\par\endgroup

  \item[Prompt 2:] \hfill \\
  \begingroup\ttfamily
 **Task Description:**  

Transform a given query into a significantly less readable version while preserving its original semantic meaning as closely as possible.

 **Constraints \& Goals:**  

- **Readability:** The rewritten text must have a **Flesch Reading Ease Score below 60**, preferably below 50.  

- **Semantic Similarity:** The rewritten text must achieve an **SBERT similarity score > 0.7** with the original query.  

- **Length Consistency:** The modified text should be **within ±10\% of the original length**.  

- **Preserve Key Terminology:** **Do not alter domain-specific words, abbreviations, or technical jargon** (e.g., "IRS," "distance").  

- **Abbreviation Handling:** **Do not expand abbreviations** unless they are already expanded in the original query.  

- **Maintain Original Intent:** Do not add, remove, or alter the factual content of the query.  

- **Retain Question Structure:** If the input is a question, the output must also be a question.  

 **Techniques to Decrease Readability:**  

1. **Lexical Complexity:** Replace common words with **advanced, academic, or formal synonyms**, while keeping domain-specific terms unchanged.  

2. **Syntactic Complexity:** Introduce **passive voice, nominalizations, embedded clauses, or subordinate structures** to increase sentence density.  

3. **Redundancy \& Formality:** Use **circumlocution, excessive formality, and indirect phrasing** (e.g., "in light of the fact that" instead of "because").  

4. **Dense Sentence Structure:** Prefer **wordy, indirect, and convoluted constructions** over direct phrasing.  

Original Query: ...

Less Readable Query:

\par\endgroup

  \item[Prompt 3:] \hfill \\
  \begingroup\ttfamily
 **Objective:**  

You are tasked with **rewriting a given query to make it significantly less readable** while preserving its original semantic meaning with high fidelity.

 **Guiding Principles:**  

- **Readability Constraint:** The rewritten text must have a **Flesch Reading Ease Score of <= 60**, preferably <= 50.  

- **Semantic Integrity:** Ensure an **SBERT similarity score of at least 0.7** between the original and rewritten text.  

- **Length Tolerance:** Maintain an **approximate length deviation of no more than ±10\%** from the original.  

- **Terminology Preservation:** Domain-specific terms (e.g., "IRS," "distance") **must remain intact** or be substituted only with **near-synonymous equivalents**.  

- **Abbreviation Handling:** If an abbreviation exists, **retain it as is** unless the original query explicitly expands it.  

- **Strict Content Preservation:** Do **not introduce any new information** or omit existing details.  

- **Question Retention:** If the input is a question, the reformulated output **must remain a question**.  

 **Techniques for Readability Reduction:**  

- **Lexical Sophistication:** Replace commonplace words with **more complex, formal, or technical alternatives** while maintaining clarity of meaning.  

- **Structural Density:** Employ **passive constructions, embedded clauses, and nominalized phrases** to increase syntactic complexity.  

- **Circumlocution \& Wordiness:** Favor **verbose, indirect expressions** over concise phrasing (e.g., "with regard to" instead of "about").  

- **Elaborate Phrasing:** Use **multi-clause structures and intricate sentence formations** to reduce direct readability.  

Original Query: ...

Less Readable Query:

\par\endgroup

\end{description}

\subsection{Politeness}

\begin{description}

  \item[Prompt 1:] \hfill \\
  \begingroup\ttfamily
 Task: Rewrite Queries to Sound More Polite and Courteous  

Rephrase the given query into a more polite, respectful, and considerate version while preserving its original intent. The output should reflect a natural, well-mannered tone suitable for professional or friendly interactions. The generated query should be a single sentence.

 Critical Rules:  

- Use a variety of politeness techniques, including warm greetings, indirect requests, and expressions of gratitude.  

- Avoid robotic or overly formal constructions—make it sound naturally courteous, warm and friendly.  

- Do not always start your sentence with 'Could you please tell'. Use emotional undertones and specific attempts at politeness.

- Maintain the original meaning without unnecessary embellishment.

- Do not start the generated query with 'I hope you are ...' or end with a single 'Thank you' sentence. Generate only a single polite query sentence.

Original Query: ...

Polite Query:
  \par\endgroup

  \item[Prompt 2:] \hfill \\
  \begingroup\ttfamily
 Task: Enhance the Courtesy of a Given Query
    
Transform the provided query into a more respectful, friendly, and warm version, ensuring it conveys respect and warmth while keeping the original intent intact. The reworded request should sound engaging, professional, and well-mannered. The generated query should be a single sentence.

 Key Considerations:

- Use a mix of politeness techniques, including indirect phrasing, friendly introductions, and appreciative language.

- Keep the tone natural—avoid overly rigid or formal wording that feels robotic.

- Vary sentence structures instead of defaulting to "Could you please...". Use emotional undertones and specific attempts at politeness.

- Maintain the original meaning while subtly enhancing the request's politeness and friendliness.

- Avoid beginning the generated query with 'I hope you are...' or concluding it with a separate 'Thank you.' sentence. Generate only one polite query sentence.

Original Query: ...

Polite Query:
\par\endgroup

  \item[Prompt 3:] \hfill \\
  \begingroup\ttfamily
 Task: Refining Queries for Politeness and Warmth

Transform a given query into a more courteous, engaging, and warm request while ensuring it retains the original intent. The revised version should sound friendly, professional, and respectful. The generated query should be a single sentence.

 Guidelines:

- Incorporate politeness techniques such as indirect requests, warm introductions, and appreciative language.

- Ensure the tone is natural—avoid excessive formality that feels robotic.

- Diversify sentence structures rather than defaulting to "Could you please...". Use emotional undertones and specific attempts at politeness.

- Subtly enhance warmth and professionalism while preserving clarity and intent.

- Avoid beginning the generated query with 'I hope you are ...' or concluding it with a standalone 'Thank you' sentence. Generate only one polite query sentence.  

Original Query: ...

Polite Query:
\par\endgroup

\end{description}

\section{LLMs Prompts}\label{app:llm_prompts}

\subsection{Few-shot Prompts}

\begin{description}
\item[Few-shot examples:] \hfill \\
  \begingroup\ttfamily

  \textbf{PopQA}: {
  
    - readability: 
    
    Question: What genre is Golden?
    
    Answer: rock music
    
    Question: In which specific genre does the work titled "Golden" find its classification?
    
    Answer: rock music
    
    - politeness: 
    
    Question: What genre is Golden?
    
    Answer: rock music
    
    Question: Would you be so kind as to share with me what genre Golden falls under?
    
    Answer: rock music

    - formality: 
    
    Question: What genre is Golden?
    
    Answer: rock music
    
    Question: Hey, so like, do you know what genre Golden is?
    
    Answer: rock music
    
    - round-trip translation: 
    
    Question: What genre is Golden?
    
    Answer: rock music
    
    Question: What genre of Golden?
    
    Answer: rock music
    
    - typos: 
    
    Question: What genre is Golden?
    
    Answer: rock music
    
    Question: What genra is Golden?
    
    Answer: rock music
}

\textbf{EntityQuestions}: {

    - readability
    
    Question: Where was Michael Jack born?
    
    Answer: Folkestone
    
    Question: In what geographical locale did the individual known as Michael Jackson enter into existence?
    
    Answer: Folkestone

    - politeness: 
    
    Question: Where was Michael Jack born?
    
    Answer: Folkestone
    
    Question: Would you be so kind as to share the birthplace of Michael Jack?
    
    Answer: Folkestone
    
    - formality: 
    
    Question: Where was Michael Jack born?
    
    Answer: Folkestone
    
    Question: Hey, so like, do you know where Michael Jack was born?
    
    Answer: Folkestone
    
    - round-trip translation: 
    
    Question: Where was Michael Jack born?
    
    Answer: Folkestone
    
    Question: Where was Michael Jacques born?
    
    Answer: Folkestone
    
    - typos: 
    
    Question: Where was Michael Jack born?
    
    Answer: Folkestone
    
    Question: Where was Michael Jack born?
    
    Answer: Folkestone
}

\textbf{MS MARCO}: {

    - readability: 
    
    Question: how long can chicken stay good in the fridge
    
    Answer: 1 to 2 days
    
    Question: What is the time span within which chicken can sustain its quality for consumption when preserved in a refrigerated setting?
    
    Answer: 1 to 2 days

    - politeness: 
    
    Question: how long can chicken stay good in the fridge
    
    Answer: 1 to 2 days
    
    Question: Would you be so kind as to share how long chicken remains fresh in the refrigerator?
    
    Answer: 1 to 2 days
    
    - formality: 
    
    Question: how long can chicken stay good in the fridge
    
    Answer: 1 to 2 days
    
    Question: Hey, so like, do you know how long chicken can last in the fridge?
    
    Answer: 1 to 2 days
    
    - round-trip translation: 
    
    Question: how long can chicken stay good in the fridge
    
    Answer: 1 to 2 days
    
    Question: How long will chicken stay fresh in the refrigerator
    
    Answer: 1 to 2 days
    
    - typos: 
    
    Question: how long can chicken stay good in the fridge
    
    Answer: 1 to 2 days
    
    Question: how leng can chickon stay good in the fridge
    
    Answer: 1 to 2 days
}

\textbf{Natural Questions}: {
    
    - readability: 
    
    Question: how many pieces in a terry's chocolate orange
    
    Answer: six
    
    Question: What is the total quantity of individual segments contained within a Terry's chocolate orange confectionery item?
    
    Answer: six
    
    - politeness: 
    
    Question: how many pieces in a terry's chocolate orange
    
    Answer: six
    
    Question: Would you be so kind as to share the number of segments typically found in a Terry's chocolate orange?
    
    Answer: six
    
    - formality: 
    
    Question: how many pieces in a terry's chocolate orange
    
    Answer: six
    
    Question: Hey, so like, do you know a terry's chocolate orange contains how many pieces
    
    Answer: six
    
    - round-trip translation: 
    
    Question: how many pieces in a terry's chocolate orange
    
    Answer: six
    
    Question: How many pieces of Terry's Chocolate Orange
    
    Answer: six
    
    - typos: 
    
    Question: how many pieces in a terry's chocolate orange
    
    Answer: six
    
    Question: how meny pieces in a tarry's chocolate orange
    
    Answer: six
}
  \par\endgroup

  \item[Prompt:] \hfill \\
  \begingroup\ttfamily
You are a professional question-answer task assistant. Use the following pieces of retrieved context to answer the question briefly. 

Context: 

{contexts}

Below are examples of questions and answers:

{few\_shot\_examples}

Now, it's your turn to answer the question below. The answer should contain ONLY one sentence and DO NOT explain reasons.
  \par\endgroup
\end{description}

\section{Rewriting Examples}

\begin{table*}[!htp]\centering
\resizebox{\textwidth}{!}{
\begin{tabular}{llp{0.85\textwidth}}\toprule
Category & Type & Query \\\midrule

\multirow{2}{*}{RTT} & Original & What type of music does The Eruption of Mount St. Helens! play? \\
                    & Rewritten & What music the eruption of Mount St Helens! play? \\
\midrule
\multirow{2}{*}{RTT} & Original & Who is Hilde Coppi married to? \\
                    & Rewritten & With whom was Hilde Coppi married? \\
\midrule

\multirow{2}{*}{RTT} & Original & Which company is HMS Blankney produced by? \\
                    & Rewritten & What company is producing HMS Blankey? \\
\midrule

\multirow{2}{*}{RTT} & Original & Where is Flemington Racecourse located? \\
                    & Rewritten & Where is Flemington Racecourse? \\
\midrule

\multirow{2}{*}{Typos} & Original & Where was R. Kent Greenawalt born? \\
                      & Rewritten & Wher was R. Kent Greenawalt born? \\
\midrule

\multirow{2}{*}{Typos} & Original & What kind of work does M. Ramanathan do? \\
                      & Rewritten & What kind ofh work does M. Ramanathan do? \\
\midrule

\multirow{2}{*}{Typos} & Original & What type of music does El Cantor del circo play? \\
                      & Rewritten & What type of music does El Cantorh del circo pay? \\
\midrule

\multirow{2}{*}{Typos} & Original & What is Rembrandt famous for? \\
                      & Rewritten & What is Rembrandt faumous for? \\
\midrule

\multirow{2}{*}{Formality} & Original & Which company is Galaxy Camera produced by? \\
                          & Rewritten & Hey, quick question! Which company actually makes the Galaxy Camera? \\
\midrule

\multirow{2}{*}{Formality} & Original & Who is the author of Intensity? \\
                          & Rewritten & Yo, do you know who wrote Intensity? \\
\midrule

\multirow{2}{*}{Formality} & Original & Which country is Parchliny located in? \\
                          & Rewritten & Hey, just curious, do you know what country Parchliny is in? \\
\midrule

\multirow{2}{*}{Formality} & Original & Who is Liu Bei's child? \\
                          & Rewritten & Hey, so, do you know who Liu Bei's kid is? I'm super curious about it! \\
\midrule

\multirow{2}{*}{Readability} & Original & What type of music does Anbe Sivam play? \\
                            & Rewritten & What genre of musical compositions is performed by Anbe Sivam? \\
\midrule

\multirow{2}{*}{Readability} & Original & Where was FC Utrecht founded? \\
                            & Rewritten & In what location was the establishment of FC Utrecht initiated? \\
\midrule

\multirow{2}{*}{Readability} & Original & Where was John Ernle educated? \\
                            & Rewritten & At which institution did John Ernle receive his education? \\
\midrule

\multirow{2}{*}{Readability} & Original & Where was The Shiru Group founded? \\
                            & Rewritten & In which geographical location did The Shiru Group originate? \\
\midrule

\multirow{2}{*}{Politeness} & Original & What music label is Time in Place represented by? \\
                           & Rewritten & May I kindly inquire which music label represents Time in Place? \\
\midrule

\multirow{2}{*}{Politeness} & Original & Which country was The Border Blasters created in? \\
                           & Rewritten & Would you be so kind as to share which country The Border Blasters originated from? \\
\midrule

\multirow{2}{*}{Politeness} & Original & Which country is Oleksin, Otwock County located in? \\
                           & Rewritten & Could you kindly share which country Oleksin, Otwock County is situated in? \\
\midrule

\multirow{2}{*}{Politeness} & Original & Where did Wolfe Tone die? \\
                           & Rewritten & Would you be so kind as to share the location where Wolfe Tone passed away? \\

\bottomrule
\end{tabular}}
\caption{Rewriting examples across all linguistic variations from the EntityQuestions dataset, with queries split across rows for readability.}
\label{tab:rew_ex}
\end{table*}

\clearpage
\section{Computational Resources}
Our experimental setup utilized models of varying scales: Gemma-2 (2B, 9B, 27B parameters), Llama-3.1 (8B, 70B parameters), and Qwen-2.5 (3B, 7B, 32B, 72B parameters). For retrieval, we employed ModernBERT Embed (149M parameters) and Contriever.
We conducted comprehensive evaluations across 5 linguistic dimensions (4 dimensions plus 2 grammatical correctness subtypes), 4 datasets, 9 language models, and 2 retrieval systems, totaling 360 experimental configurations. Each model inference run required approximately 1.5 hours, resulting in 540 GPU hours on 16 L40S GPUs distributed across different model configurations. Retrieval evaluation required an additional 40 GPU hours.
Data rewriting was performed using GPT-4o-mini, requiring 40 hours of API usage. Total computational cost comprised 620 GPU hours on L40S hardware plus commercial API usage for data preprocessing.